%% file: acl_latex.tex
\pdfoutput=1

\documentclass[11pt]{article}

\usepackage[]{acl}

\usepackage{times}
\usepackage{latexsym}
\usepackage{array}
\usepackage{arydshln}
\usepackage{marvosym}
\usepackage{picins}
\usepackage{amsmath}

\usepackage[T1]{fontenc}

\usepackage[utf8]{inputenc}

\usepackage{microtype}
\usepackage{tikz}
\usepackage{pgfplots}
\usepackage[edges]{forest}
\definecolor{paired-light-blue}{RGB}{198, 219, 239}
\definecolor{paired-dark-blue}{RGB}{49, 130, 188}
\definecolor{paired-light-orange}{RGB}{251, 208, 162}
\definecolor{paired-dark-orange}{RGB}{230, 85, 12}
\definecolor{paired-light-green}{RGB}{199, 233, 193}
\definecolor{paired-dark-green}{RGB}{49, 163, 83}
\definecolor{paired-light-purple}{RGB}{218, 218, 235}
\definecolor{paired-dark-purple}{RGB}{117, 107, 176}
\definecolor{paired-light-gray}{RGB}{217, 217, 217}
\definecolor{paired-dark-gray}{RGB}{99, 99, 99}
\definecolor{paired-light-pink}{RGB}{222, 158, 214}
\definecolor{paired-dark-pink}{RGB}{123, 65, 115}
\definecolor{paired-light-red}{RGB}{231, 150, 156}
\definecolor{paired-dark-red}{RGB}{131, 60, 56}
\definecolor{paired-light-yellow}{RGB}{231, 204, 149}
\definecolor{paired-dark-yellow}{RGB}{141, 109, 49}
\tikzset{%
    parent/.style =          {align=center,text width=2.5cm,rounded corners=3pt, line width=0.3mm, fill=gray!10,draw=gray!80},
    child/.style =           {align=center,text width=2.3cm,rounded corners=3pt, fill=blue!10,draw=blue!80,line width=0.3mm},
    grandchild/.style =      {align=center,text width=2cm,rounded corners=3pt},
    greatgrandchild/.style = {align=center,text width=1.5cm,rounded corners=3pt},
    greatgrandchild2/.style = {align=center,text width=1.5cm,rounded corners=3pt},    
    referenceblock/.style =  {align=center,text width=1.5cm,rounded corners=2pt},
    acquisition/.style =    {align=center,text width=2.2cm,rounded corners=3pt, fill=paired-light-blue!50,draw=paired-dark-blue!65,line width=0.3mm},   
    acquisition_work/.style =           {align=center, text width=4.5cm,rounded corners=3pt, fill=paired-light-blue!50,draw=blue!0,line width=0.3mm},  
    representation/.style =           {align=center,text width=2.2cm,rounded corners=3pt, fill=paired-light-orange!50,draw=paired-dark-orange!65,line width=0.3mm},   
    representation_work/.style =           {align=center,text width=4.5cm,rounded corners=3pt, fill=paired-light-orange!50,draw=red!0,line width=0.3mm},    
    probing/.style =           {align=center,text width=2.2cm,rounded corners=3pt, fill= paired-light-green!50,draw=paired-dark-green!75,line width=0.3mm},   
    probing_work/.style =           {align=center,text width=4.5cm,rounded corners=3pt, fill= paired-light-green!50,draw= cyan!0,line width=0.3mm},      
    editing/.style =           {align=center,text width=2.2cm,rounded corners=3pt, fill= paired-light-purple!50,draw=paired-dark-purple!75,line width=0.3mm},   
    editing_work/.style =           {align=center,text width=4.5cm,rounded corners=3pt, fill= paired-light-purple!50,draw= orange!0,line width=0.3mm},        
    application/.style =           {align=center,text width=2.2cm,rounded corners=3pt, fill= paired-light-red!35,draw=paired-light-red!90,line width=0.3mm},   
    application_work/.style =           {align=center,text width=4.5cm,rounded corners=3pt, fill= paired-light-red!35,draw= magenta!0,line width=0.3mm},         
}

\usepackage{times}
\usepackage{latexsym}
\usepackage{microtype}

\usepackage{inconsolata}

\usepackage{soul}
\usepackage{makecell}
\usepackage{fontawesome}
\usepackage{makecell}

\usepackage{graphicx}
\usepackage{booktabs}
\usepackage{multirow}
\usepackage{multicol}
\usepackage{amssymb}
\usepackage{graphicx}

\NewDocumentCommand\emojismile{}{{\includegraphics[scale=.045]{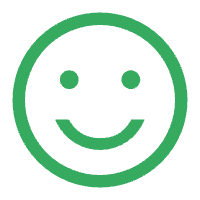}}}

\title{The Life Cycle of Knowledge in Big Language Models: A Survey}

\author{Boxi Cao${}^{1,3}$, Hongyu Lin${}^{1}$, Xianpei Han${}^{1, 2}$\textsuperscript{\Letter}, Le Sun${}^{1,2}$\\
${}^{1}$Chinese Information Processing Laboratory ~ ${}^{2}$State Key Laboratory of Computer Science \\
Institute of Software, Chinese Academy of Sciences, Beijing, China\\
${}^{3}$University of Chinese Academy of Sciences, Beijing, China \\
{\tt \{boxi2020,hongyu,xianpei,sunle\}@iscas.ac.cn}
}

\begin{document}
\maketitle
\begin{abstract}

Knowledge plays a critical role in artificial intelligence. 
Recently, the extensive success of pre-trained language models (PLMs) has raised significant attention about how knowledge can be acquired, maintained, updated and used by language models. Despite the enormous amount of related studies, there still lacks a unified view of how knowledge circulates within language models throughout the learning, tuning, and application processes, which may prevent us from further understanding the connections between current progress or realizing existing limitations. In this survey, we revisit PLMs as knowledge-based systems by dividing the life circle of knowledge in PLMs into five critical periods, and investigating how knowledge circulates when it is built, maintained and used. To this end, we systematically review existing studies of each period of the knowledge life cycle, summarize the main challenges and current limitations, and discuss future directions\footnote{We openly released a corresponding paper list which will be regularly updated on \url{https://github.com/c-box/KnowledgeLifecycle}.}.

\end{abstract}


\input{rewrite/new_intro}

\input{sections/acquisition.tex}
\input{sections/representation.tex}
\input{sections/probing.tex}
\input{sections/editing.tex}

\input{sections/application.tex}
\input{sections/conclusion.tex}

\bibliography{all}
\bibliographystyle{acl_natbib}






\end{document}

%% file: rewrite/new_intro.tex
\section{Introduction}
\label{sec:intro}

\begin{quote}
\textit{Fundamentally, AI is the science of knowledge -- how to represent knowledge and how to obtain and use knowledge.}
\end{quote}
\rightline{\citet{Nilson:ai:1974}}

Knowledge is the key to high-level intelligence. How a model obtains, stores, understands and applies knowledge has long been a critical research topic in machine intelligence. 
Recent years have witnessed the rapid development of pre-trained language models (PLMs). 
Through self-supervised pre-training on large-scale unlabeled corpora, PLMs show strong generalization and transferring abilities across different tasks/datasets/settings over previous methods, and therefore have achieved remarkable success in natural language processing~\citep{devlin2018bert, liu2019roberta, raffel2020exploring,radford2019language,brown2020language,lewis2019bart}.

\begin{figure}[!tp]
\centering
    \includegraphics[width=0.5\textwidth]{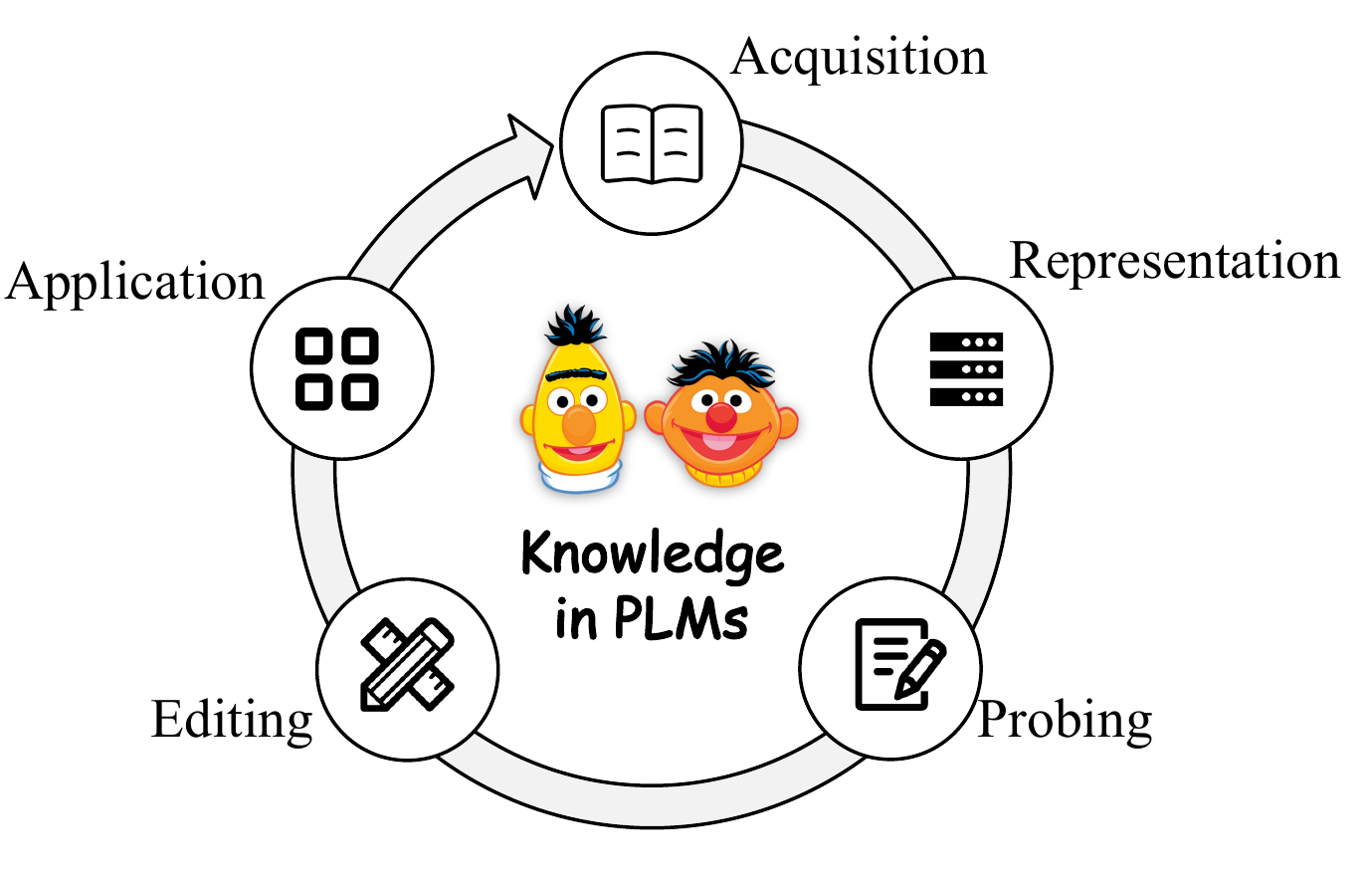}
    \caption{Five critical periods in life circle of knowledge in language models.}
    \label{fig:circle_3}
\end{figure}



The success of pre-trained language models has raised great attention about the nature of their entailed knowledge. There have been numerous studies focusing on how knowledge can be acquired, maintained, and used by pre-trained language models. 
Along these lines, many novel research directions have been explored. For example, knowledge infusing devotes to injecting explicit structured knowledge into PLMs~\citep{sun2019ernie,zhang2019ernie,sachan2020syntax}. Knowledge probing aims to evaluate the type and amount of knowledge stored in PLMs' parameters~\citep{petroniLanguageModelsKnowledge2019,linOpenSesameGetting2019,hewittStructuralProbeFinding}. And knowledge editing is dedicated to modifying the incorrect or undesirable knowledge acquired by PLMs~\citep{zhu2020modifying,de2021editing,mitchell2021fast}.


Despite the large amount of related studies, current studies primarily focus on one specific stage of knowledge process in PLMs, thereby lacking a unified perspective on how knowledge circulates throughout the entire model learning, tuning, and application phases. 
The absence of such comprehensive studies makes it hard to better understand the connections between different knowledge-based tasks, discover the correlations between different periods during the knowledge life circle in PLMs, exploit the missing links and tasks for investigating knowledge in PLMs, or explore the shortcomings and limitations of existing studies.
For example, while numerous studies attempt to assess the knowledge in language models that are already pre-trained, there are few studies dedicated to investigating why PLMs can learn from pure text without any supervision about knowledge, as well as how PLMs represent or store these knowledge.
Meanwhile, many researchers have tried to explicitly inject various kinds of structural knowledge into PLMs , but few studies propose to help PLMs better acquire specific kinds of knowledge from pure text by exploiting the knowledge acquisition mechanisms behind.
As a result, related research may be overly focused on several directions but fail to comprehensively understand, maintain and control knowledge in PLMs, and therefore limits the improvements and further application.
%

In this survey, we propose to systematically review the knowledge-related studies in pre-trained language models from a knowledge engineering perspective. Inspired by research in cognitive science~\citep{zimbardo1975psychology,churchland1988perspectives} and knowledge engineering~\citep{studer1998knowledge,schreiber2000knowledge}, we regard pre-trained language models as knowledge-based systems, and investigate the life cycle of how knowledge circulates when it is acquired, maintained and used in pre-trained models~\citep{studer1998knowledge,schreiber2000knowledge}. Specifically, we divide the life cycle of knowledge in pre-trained language models into the following five critical periods as shown in Fig.~\ref{fig:circle_3}:



\begin{itemize}
    \item \textbf{Knowledge Acquisition}, which focuses on the procedure of language models learning various knowledge from text or other knowledge sources.
    \item \textbf{Knowledge Representation}, which focuses on the underlying mechanism of how different kinds of knowledge are transformed, encoded, and distributed in PLMs' parameters.
    \item \textbf{Knowledge Probing}, which aims to evaluate how well current PLMs entailing different types of knowledge.
    \item \textbf{Knowledge Editing}, which tries to edit or delete knowledge containing in language models.
    \item \textbf{Knowledge Application}, which tries to distill or leverage knowledge in pre-trained language models for practical application.
\end{itemize}

\input{figures/typology}

For each of these periods, we sort out the existing studies, summarize the main challenges and limitations, and discuss future directions. 
Based on the unified perspective, we are able to understand and utilize the close connections between different periods instead of consider them as independent tasks.
For instance, understanding the knowledge representation mechanism of PLMs is valuable for researchers to design better knowledge acquisition objectives and knowledge editing strategies. Proposing reliable knowledge probing methods could help us find the suitable applications for PLMs, and gain insight into their limitations, thereby facilitating improvement.
Through this survey, we are willing to comprehensively conclude the progress, challenges and limitations of current studies, help researchers better understand the whole field from a novel perspective, and shed light on the future directions about how to better regulate, represent and apply the knowledge in language models from a unified perspective.

We summarize our contributions as follows:
\begin{itemize}
    \item We propose to revisit pre-trained language models as knowledge-based systems, and divide the life cycle of knowledge in PLMs into five critical periods. 
    \item For each period, we review existing studies, summarize the main challenges and shortcomings for each direction. 
    \item Based on this review, we discuss about the  limitations of the current research, and shed light to potential future directions. 
\end{itemize}

\section{Overview}

In this section, we present the overall structure of this survey, describe our taxonomy shown in Fig.~\ref{fig:tax} in detail, and discuss the topics in each critical period. 

\textbf{Knowledge Acquisition} is the knowledge learning procedure of language models.
Currently, there are two main sources for knowledge acquisition: the plain text data and the structured data.
For acquiring knowledge from text data, LMs typically conduct self-supervised learning on large-scale text corpora~\citep{devlin2018bert, liu2019roberta, brown2020language,raffel2020exploring}.
This survey will focus on the methods and mechanisms of how pre-trained language models obtaining knowledge from pure texts~\citep{chiang2020pretrained,perez2021much,liu-etal-2021-probing-across}.
For acquiring knowledge from structured data, current research focus on knowledge injection from different kinds of structured data into PLMs. 
The primary categories of structured data contains entity knowledge~\citep{sun2019ernie, xiong2019pretrained,peters2019knowledge}, factual knowledge~\citep{zhang2019ernie, wang2021kepler, wang2020k, liu2020k}, commonsense knowledge~\citep{bosselut2019comet,ye2019align,guan2020knowledge,ma2021knowledge} and linguistic knowledge~\citep{ke2019sentilare,lauscher2019specializing,zhou2019limit,bai2021syntax}. 
We will discuss all of them in Section~\ref{sec:acquire}.

\textbf{Knowledge Representation} aims to investigate how language models encode, store and represent knowledge in their dense parameters. 
The investigation about the knowledge representation mechanisms will aid in a better understanding and control of knowledge in PLMs, and may also inspire researchers for better understanding the knowledge representation in human brains.
Currently, the strategies for knowledge representation analysis in PLMs include gradient-based~\citep{geva2020transformer,dai2021knowledge}, causal-inspired~\citep{meng2022locating}, attention-based~\citep{DBLP:conf/blackboxnlp/ClarkKLM19,htutAttentionHeadsBERT2019,linOpenSesameGetting2019}, 
and layer-wise~\citep{linOpenSesameGetting2019, liu-etal-2019-linguistic, juneja2022finding} methods. We will discuss them in Section~\ref{sec:repre}.

\textbf{Knowledge Probing} aims to evaluate how well current PLMs entailing specific types of knowledge.
Currently, two primary strategies are used to probe the knowledge in PLMs:
1) Prompt-based probing, which usually constructs knowledge-instructed prompt, then query PLMs using these natural language expressions~\citep{petroniLanguageModelsKnowledge2019,jiangXFACTRMultilingualFactual2020,sung2021can, forbesNeuralLanguageRepresentations2019,Zhou2020EvaluatingCI}. For example, querying PLMs with ``The capital of France is \underline{\hbox to 4mm{}}.'' to evaluate whether PLMs have stored the corresponding knowledge <France, capital, Paris>. Meanwhile, to improve PLMs' performance, a series of studies devote to optimizing prompts in both discrete~\citep{jiangHowCanWe2020,feldmanCommonsenseKnowledgeMining2019,haviv-etal-2021-bertese,shinAutoPromptElicitingKnowledge2020} and continual space~\citep{zhong2021factual,liPrefixTuningOptimizingContinuous2021,liu2021gpt}. 
Despite the widely application of prompt-based probing, lots of studies also point out that there still exist some pending issues such as inconsistent~\citep{elazar2021measuring,kassnerNegatedMisprimedProbes2020b,jang2022can,cao-etal-2022-prompt}, inaccurate~\citep{poernerEBERTEfficientYetEffectiveEntity2020,zhong2021factual,cao-etal-2021-knowledgeable} and unreliable~\citep{cao-etal-2021-knowledgeable,li-etal-2022-pre}, and question the quantity results of prompt-based probing.
2) Feature-based probing, which normally freezes the parameters of original PLMs, and evaluates PLMs on probing tasks based on their internal representation or attention weights. 
We categorize existing feature-based probing studies into classifier-based probing~\citep{linOpenSesameGetting2019,tenneyWhatYouLearn2019,DBLP:conf/blackboxnlp/ClarkKLM19,liu-etal-2019-linguistic} and classifier-free probing~\citep{wu2020perturbed,zhou2021directprobe} according to whether an additional classifier is introduced.
Since most methods introduce additional parameters or training data, the main shortcoming of feature-based probing is whether the results should attribute to knowledge in PLMs or probing task learned by additional probes.
We will discuss them in Section~\ref{sec:prob}.

\textbf{Knowledge Editing} aims to modify the incorrect knowledge or delete the undesirable information in PLMs.
Because of inevitable mistakes learned by PLMs and the update of knowledge, reliable and effective knowledge editing approaches are essential for the sustainable application of PLMs. 
Current approaches include constrained fine-tuning~\citep{zhu2020modifying}, memory-based~\citep{mitchell2022memory,madaan2022memory,dong2022calibrating}, meta-learning inspired~\citep{de2021editing,hase2021language,mitchell2021fast} and location-based methods~\citep{dai2021knowledge,meng2022locating}. 
We will discuss them in Section~\ref{sec:edit}.

\textbf{Knowledge Application} aims to distill or leverage specific knowledge from PLMs to benefit further applications. Currently, there are two main kinds of application paradigms for knowledge in PLMs:
1) Language models as knowledge bases (LMs-as-KBs), which regards language models as dense knowledge bases that can be directly queried with natural language to obtain specific types of knowledge~\citep{petroniLanguageModelsKnowledge2019,heinzerling2020language,jiangHowCanWe2020,wang2020language,cao-etal-2021-knowledgeable,razniewski2021language,alkhamissi2022review}. 
And we provide a comprehensive comparison between structured knowledge bases and LMs-as-KBs~\citep{razniewski2021language} from four aspects, including construction, coverage, interaction and reliability; 
2) Language models for downstream task, which directly uses PLMs entailing specific kinds of knowledge in downstream NLP tasks via fine-tuning~\citep{manning2020emergent,wei2021knowledge,yang2021survey,yin2022survey}, prompt-learning~\citep{radfordLanguageModelsAre2019,brown2020language,liu2021pre} and in-context learning~\citep{brown2020language,zhao2021calibrate,lu2021fantastically}. 
We will discuss them in Section~\ref{sec:application}.

%% file: figures/typology.tex
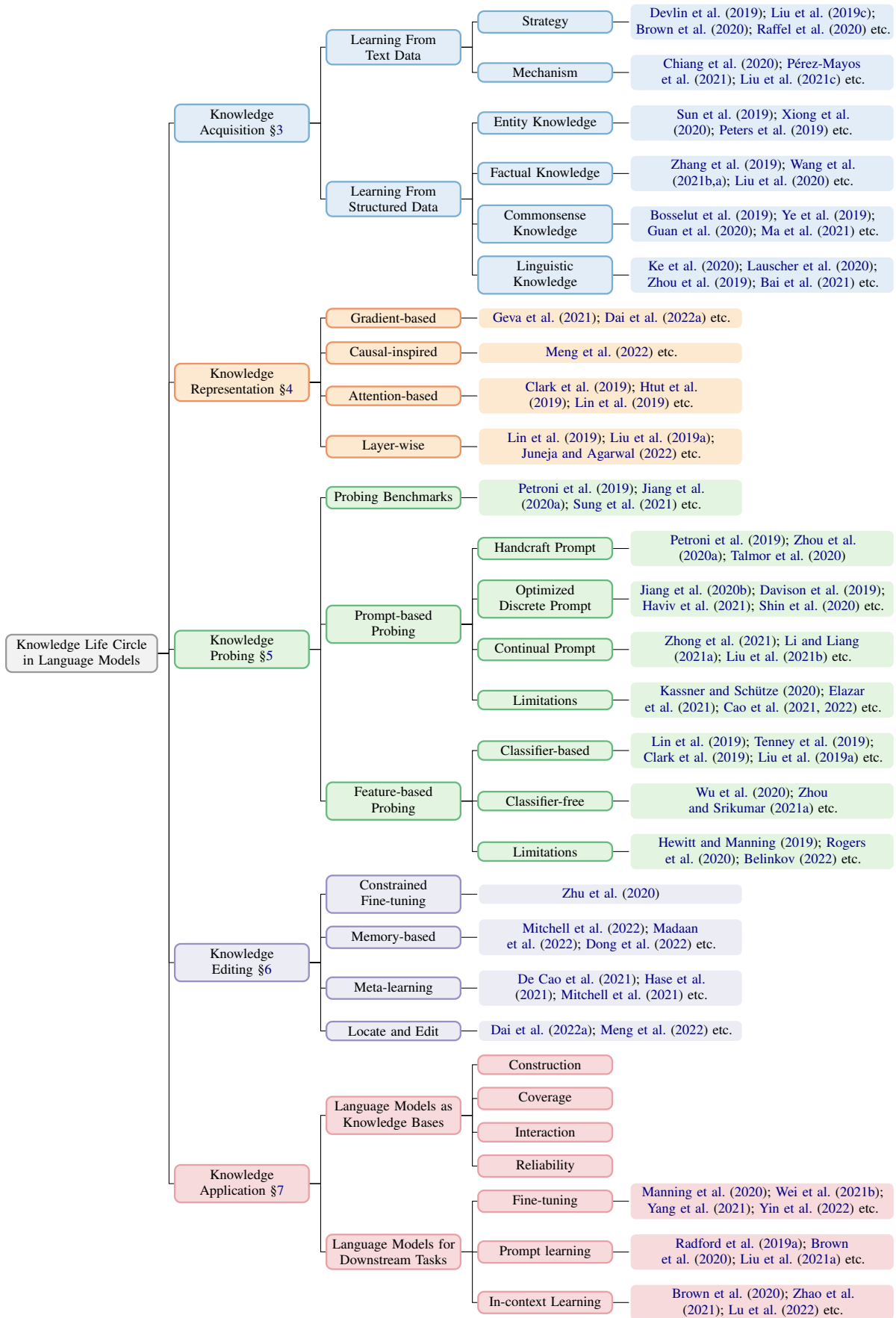
\begin{figure*}[!ht]
\scriptsize
    \begin{forest}
        for tree={
            forked edges,
            grow'=0,
            draw,
            rounded corners,
            node options={align=center,},
            text width=2.7cm,
            s sep=6pt,
            calign=edge midpoint,
        },
        [Knowledge Life Circle in Language Models, fill=gray!45, parent
            [Knowledge\\ Acquisition \S\ref{sec:acquire}, for tree={acquisition}
                [Learning From Text Data,  acquisition
                    [Strategy, acquisition
                        [\citet{devlin2018bert, liu2019roberta, brown2020language,raffel2020exploring} etc., acquisition_work]
                    ]
                    [Mechanism, acquisition
                        [\citet{chiang2020pretrained,perez2021much,liu-etal-2021-probing-across} etc., acquisition_work]
                    ]
                ]
                [Learning From Structured Data,  acquisition
                    [Entity Knowledge,  acquisition
                        [\citet{sun2019ernie, xiong2019pretrained,peters2019knowledge} etc., acquisition_work]
                    ]
                    [Factual Knowledge,  acquisition
                        [\citet{zhang2019ernie, wang2021kepler, wang2020k, liu2020k} etc., acquisition_work]
                    ]
                    [Commonsense Knowledge, acquisition
                        [\citet{bosselut2019comet,ye2019align,guan2020knowledge,ma2021knowledge} etc., acquisition_work]
                    ]
                    [Linguistic Knowledge, acquisition
                        [\citet{ke2019sentilare,lauscher2019specializing,zhou2019limit,bai2021syntax} etc., acquisition_work]
                    ]
                ]
            ]
            [Knowledge\\ Representation \S\ref{sec:repre}, for tree={representation}
                [Gradient-based, representation
                    [\citet{geva2020transformer,dai2021knowledge} etc., representation_work]
                ]
                [Causal-inspired, representation
                    [\citet{meng2022locating} etc., representation_work]
                ]
                [Attention-based, representation
                    [\citet{DBLP:conf/blackboxnlp/ClarkKLM19,htutAttentionHeadsBERT2019,linOpenSesameGetting2019} etc., representation_work]
                ]
                [Layer-wise, representation
                    [\citet{linOpenSesameGetting2019, liu-etal-2019-linguistic, juneja2022finding} etc., representation_work]
                ]
            ]
            [Knowledge\\ Probing \S\ref{sec:prob}, for tree={probing}
                [Probing Benchmarks, probing
                    [\citet{petroniLanguageModelsKnowledge2019,jiangXFACTRMultilingualFactual2020,sung2021can} etc., probing_work]
                ]
                [Prompt-based Probing, probing
                    [Handcraft Prompt, probing
                        [\citet{petroniLanguageModelsKnowledge2019,Zhou2020EvaluatingCI,Talmor2020oLMpicsOnWL}, probing_work]
                    ]
                    [Optimized \\ Discrete Prompt, probing
                        [\citet{jiangHowCanWe2020,feldmanCommonsenseKnowledgeMining2019,haviv-etal-2021-bertese,shinAutoPromptElicitingKnowledge2020} etc., probing_work]
                    ]
                    [Continual Prompt, probing
                        [\citet{zhong2021factual,liPrefixTuningOptimizingContinuous2021,liu2021gpt} etc., probing_work]
                    ]
                    [Limitations, probing
                        [\citet{kassnerNegatedMisprimedProbes2020b,elazar2021measuring,cao-etal-2021-knowledgeable, cao-etal-2022-prompt} etc., probing_work]
                    ]
                ]
                [Feature-based Probing, probing
                    [Classifier-based, probing
                        [\citet{linOpenSesameGetting2019,tenneyWhatYouLearn2019,DBLP:conf/blackboxnlp/ClarkKLM19,liu-etal-2019-linguistic} etc., probing_work]
                    ]
                    [Classifier-free, probing
                        [\citet{wu2020perturbed,zhou2021directprobe} etc., probing_work]
                    ]
                    [Limitations, probing
                        [\citet{hewittStructuralProbeFinding,rogers2020primer,belinkov2022probing} etc., probing_work]
                    ]
                ]
            ]
            [Knowledge\\ Editing \S\ref{sec:edit}, editing
                [Constrained Fine-tuning, editing
                    [\citet{zhu2020modifying}, editing_work]
                ]
                [Memory-based, editing
                    [\citet{mitchell2022memory,madaan2022memory,dong2022calibrating} etc., editing_work]
                ]
                [Meta-learning, editing
                    [\citet{de2021editing,hase2021language,mitchell2021fast} etc., editing_work]
                ]
                [Locate and Edit, editing
                    [\citet{dai2021knowledge,meng2022locating} etc., editing_work]
                ]
            ]
            [Knowledge\\ Application \S\ref{sec:application}, application
                [Language Models as Knowledge Bases, application
                    [Construction, application]
                    [Coverage, application]
                    [Interaction, application]
                    [Reliability, application]
                ]
                [Language Models for Downstream Tasks, application
                    [Fine-tuning, application
                        [\citet{manning2020emergent,wei2021knowledge,yang2021survey,yin2022survey} etc., application_work]
                    ]
                    [Prompt learning, application
                        [\citet{radfordLanguageModelsAre2019,brown2020language,liu2021pre} etc., application_work]
                    ]
                    [In-context Learning, application
                        [\citet{brown2020language,zhao2021calibrate,lu2021fantastically} etc., application_work]
                    ]
                ]
            ]
        ]
    \end{forest}
    \caption{Typology of knowledge life circle in big language models.}
    \label{fig:tax}
\end{figure*}

%% file: sections/acquisition.tex
\section{Knowledge Acquisition}
\label{sec:acquire}

During the knowledge acquisition period, pre-trained language models learn knowledge from different knowledge sources. 
In this section, we categorize and describe knowledge acquisition strategies according to knowledge sources, and then discuss the future directions.



\subsection{Learning from Text Data}
\label{sssec:from_text}


Currently, pre-trained language models usually acquire various knowledge from pure text through self-supervised learning on a large-scale text corpus. 
In this section, we will first introduce several widely used learning objectives~\citep{xipengqiuPretrainedModelsNatural2020}, and then discuss the learning mechanisms behind them.

\textbf{Causal Language modeling} aims to autoregressively predict the next token in the input sequence, which is the most popular pre-training tasks~\citep{radford2019language,brown2020language,ouyang2022training,scao2022bloom} and has demonstrated excellent effectiveness in capturing context dependency and text generation paradigms.
One limitation of causal language modeling is unidirectional, which can only capture contextual information from left to right.

\textbf{Masked Language Modeling} aims to mask some tokens in the input randomly, and then predict the masked token conditioned on the rest of sequence~\citep{devlin2018bert, liu2019roberta}. 
Unlike causal language modeling, which can only obtain information in a unidirectional manner, masked language modeling can capture contextual information from both left-to-right and right-to-left directions.

\textbf{Seq2seq Masked Language Modeling} uses an encoder-decoder architecture for pre-training, which first feeds the encoder with masked sequence, and the decoder is supposed to predict the masked tokens autoregressively~\citep{raffel2020exploring,song2019mass}.

\textbf{Denoising Autoencoder} first corrupts the input sequence with randomly mask symbols, then feed the input into a bidirectional encoder, and the likelihood of the whole original input is calculated with an auto-regressive decoder~\citep{lewis2019bart}.


Although PLMs are pre-trained without any supervision from  external knowledge sources, they have been shown to capture a diverse range of knowledge within their parameters, such as linguistic knowledge~\citep{linOpenSesameGetting2019,tenneyWhatYouLearn2019,liuLinguisticKnowledgeTransferability2019,htutAttentionHeadsBERT2019,hewittStructuralProbeFinding,goldbergAssessingBERTSyntactic2019,warstadtInvestigatingBERTKnowledge2019}, semantic knowledge~\citep{tenneyWhatYouLearn2019,wallaceNLPModelsKnow2019,ettingerWhatBERTNot2020} and world knowledge~\citep{feldmanCommonsenseKnowledgeMining2019,bouraouiInducingRelationalKnowledge2019,forbesNeuralLanguageRepresentations2019,zhouEvaluatingCommonsensePretrained2019,robertsHowMuchKnowledge2020a,linBirdsHaveFour2020,tamborrinoPretrainingAlmostAll2020a}. 

Intuitively, PLMs learn such knowledge because they can abstract, generalize and store the implicit knowledge in the text through self-supervised learning. 
Unfortunately, the underlying mechanism of how and why PLMs acquire or forget knowledge still remains to be explored. And it will be valuable to understand the behaviors of PLMs and inspire better knowledge acquisition strategies. 



To understand the underlying mechanisms, some studies dive into the dynamics of LMs' pre-training procedure. 
Many researchers study the training dynamics of neural networks.
For example, ~\citet{achille2018critical} try to figure out whether there exist critical periods in the learning process of neural networks. 
~\citet{liu-etal-2021-probing-across} devote to finding a mathematical solution for the semantic development in deep linear networks. 
Other studies~\citep{saphra2018understanding,saphra-lopez-2020-lstms} analyze the training dynamics of LSTM~\citep{hochreiter1997long} with techniques such as SVCCA~\citep{raghu2017svcca}.

While most existing studies focus on neural networks with relatively simple architectures.
Only a few studies consider knowledge in large-scale pre-trained language models.
~\citet{chiang2020pretrained} first systematically investigate the knowledge acquisition process during the training of ALBERT~\citep{lan2019albert}. Specifically, they study the syntactic knowledge, semantic knowledge, and world knowledge development during pre-training, and find that the learning process varies across knowledge, and having more pre-trained steps could not necessarily increase the knowledge in PLMs.
~\citet{perez2021much} investigate the effect of the size of the pre-trained corpus on the syntactic ability of the RoBERTa~\citep{liu2019roberta} model, and find that models pre-trained on more data typically contain more syntactic knowledge and perform better in related downstream tasks.
~\citet{liu-etal-2021-probing-across} also investigate the knowledge acquisition process of RoBERTa~\citep{liu2019roberta} on various knowledge. And find that compared with linguistic knowledge which can be learned quickly and robustly, world knowledge is learned slowly and domain-sensitive. 

\subsection{Learning from Structured Data}

Apart from acquiring knowledge from pure text, 
PLMs can also acquire knowledge by injecting explicit structured knowledge into them.
In this section, we review these studies according to the category of structured knowledge sources.

\paragraph{Entity Knowledge} 
To learn entity knowledge explicitly, lots of studies propose entity-guided tasks for language model pretraining.
For example,
~\citet{sun2019ernie} and \citet{shen2020exploiting} use entity-level masking to enhance language models, which first recognize named entities in a sentence, then all the tokens corresponding to these entities as masked and predicted at once.
~\citet{xiong2019pretrained} present replaced entity detection, which randomly replaces the named entities in a sentence with another mention of the same entity or other entities of the same type, and LMs are supposed to determine which entities are replaced. 
~\citet{yamada2020luke} treat words and entities as independent tokens, and conduct mask language modeling separately to learn both contextualized word representation and entity representation.
~\citet{fevry2020entities} combine the mention detection and entity linking pre-training objectives with mask language modeling to match the entities in text with specific entity memories.
In addition to the entity mentions themselves, researchers have also introduced other meta-information such as entity description to further assist the entity knowledge learning~\citep{logeswaran2019zero,gillick2019learning}. 
Another efficient way to enrich PLMs' text representation with entity knowledge is utilizing word-to-entity attention~\citep{peters2019knowledge,yamada2020luke}.


\paragraph{Factual Knowledge}
In structured knowledge bases, factual knowledge is generally represented as triples (subject entity, relation, object entity). 
For a long time, researchers have been dedicated to aiding PLMs to acquire more factual knowledge to perform better on downstream tasks.
On the one hand, introducing knowledge graph embedding into the pre-training procedure could be effective. 
~\citet{zhang2019ernie} propose an aggregator to combine the corresponding knowledge embedding of the entities in text and token embedding. 
~\citet{wang2021kepler} co-train mask language modeling and knowledge graph embedding objectives, which could produce both informative text and knowledge embedding.
On the other hand, some studies propose designing factual knowledge-guided auxiliary tasks.
~\citet{wang2020k} add an adapter to infuse knowledge into PLMs without updating the original parameters. The adapter is trained with predication prediction to determine the relation type between tokens.
~\citet{qin2020erica} propose the entity discrimination tasks to predict the object entity given subject entity and relation, as well as relation discrimination tasks to predict the semantic connection between relation pairs.
~\citet{banerjee2020self} directly pre-train language model on the knowledge graph, the model is given two elements of a knowledge triple to predict the rest one.
~\citet{liu2020k} argue that incorporating a whole knowledge base into PLMs might induce the knowledge noise issue, and propose to learn from a specific sub-graph related to each input sentence.
Moreover, ~\citet{baldini-soares-etal-2019-matching} propose to learn relational knowledge solely from entity-link text through "matching in the blank" objective, which first replaces the entities in text with blank symbols and then brings the relation representations closer when they have the same pair of entities.




\paragraph{Commonsense Knowledge}
One of the most common strategy for PLMs learning commonsense knowledge is converting the knowledge to natural language expressions before learning. 
~\citet{bosselut2019comet, guan2020knowledge,shwartz2020unsupervised} first transfer the commonsense knowledge triples to natural language with prompt, then pre-train LMs on these knowledge-augmented data.
~\citet{ye2019align} post-training LMs on commonsense QA datasets created by AWS (align, mask, select).
~\citet{ma2021knowledge} transform structured commonsense knowledge into natural language questions for model learning.

\paragraph{Linguistic Knowledge}
By designing the corresponding pre-training tasks, PLMs could also learn linguistic knowledge explicitly, such as sentiment knowledge~\citep{ke2019sentilare,tian2020skep}, lexical knowledge~\citep{lauscher2019specializing,levine2019sensebert,zhou2019limit}, syntax knowledge~\citep{zhou2019limit,sachan2020syntax,bai2021syntax}, etc.
For example, to equip LMs with sentiment knowledge, ~\citet{ke2019sentilare} first label each word with a POS tag and sentiment polarity, and then incorporate both the word-level and sentence-level sentiment label with the mask language modeling objective.
Similarly, ~\citet{tian2020skep} first mine sentiment knowledge from unlabeled data based on pointwise mutual information (PMI), and then conduct pre-training tasks such as sentiment masking, sentiment word prediction and word polarity prediction with these sentiment information.
As for lexical knowledge, ~\citet{lauscher2019specializing} first acquire word similarity information from WordNet~\citep{miller1995wordnet} and BabelNet~\citep{navigli2010babelnet}, and then add word relation classification tasks in addition to BERT's original pre-training tasks.
~\citet{levine2019sensebert} also introduces the lexical information from WordNet and adds a masked-word prediction task.
To incorporate dependency knowledge with PLMs, \citet{song2022improving} construct a dependency matrix for attention alignment calibration and a fusion module to integrate dependency information.
Explicitly learning syntax knowledge also raises the researchers' attention, ~\citet{sachan2020syntax} investigate infusing syntax knowledge by either adding a syntax-GNN on the output of transformers or incorporating with text embedding using attention.
To further capture the syntax knowledge, ~\citet{bai2021syntax} using multiple attention networks, with each one encoding one relation from the syntax tree.









\subsection{Discussions and Future Work}

As we mentioned above, there have been extensive studies for better knowledge acquisition of language models, and most of them focus on infusing existing structured knowledge sources into PLMs.
The learning from text data methods can be easily scaled, and the knowledge sources is easily obtained. But the underlying mechanism is still mostly unclear, the knowledge acquisition process is implicitly and thus is hard to control, and may lead to inconsistent prediction, undesirable bias and unforeseen risks.
The learning from structured data methods can explicitly inject knowledge into PLMs, but are limited by the cost, domain, scale and quality of knowledge sources. Furthermore, since the knowledge injection methods are often specialized to specific kinds of knowledge, it is often difficult to extend or produce new knowledge.

Furthermore, because all knowledge in PLMs are implicitly encoded as parameters, it is often very difficult to control and validate the knowledge acquisition process.
There are also several studies such as retrieval-based PLMs, focus on retrieving related knowledge or context to enhance original PLMs~\citep{DBLP:journals/corr/abs-2002-08909,DBLP:conf/nips/LewisPPPKGKLYR020,DBLP:journals/corr/abs-2211-12561}, rather than injecting knowledge into PLMs' parameters.

Several future directions of knowledge acquisition in PLMs may lie in: 1) For the knowledge acquisition from existing  structured knowledge sources, it is critical to develop universal knowledge injection methods which can uniformly injecting different types of knowledge from different knowledge sources, and ensures continuous learning and avoid catastrophic forgetting in the meantime.
2) For the knowledge acquisition from pure text data, it is helpful to fully understand the underlying mechanism of knowledge learning in PLMs, and develop effective knowledge learning algorithms which can learn specific knowledge from text data in a controllable and predicable way.
3) Furthermore, it is also important to build comprehensive benchmarks for investigating and assessing the knowledge acquisition process of PLMs.

%% file: sections/representation.tex
\section{Knowledge Representation}
\label{sec:repre}

Knowledge representation studies investigate how pre-trained language models encode, transform and store the acquired knowledge. 
In PLMs, knowledge is encoded to dense vector representations and held in their distributed parameters, but how each kind of knowledge is encoded, transformed, and stored into the parameters is still unclear and needs further investigation.
Currently, a few studies have investigated the knowledge representation in language models, and we will first review these studies according to their analysis techniques.



\subsection{Analyzing Knowledge Representations in PLMs}
\label{rep_method}
Currently, the analyzing approaches for knowledge representation in PLMs can be classified into four categories: gradient-based, causal-inspired, attention-based and layer-wise methods. 
The first three methods aim to locate specific knowledge in PLMs' corresponding neurons or attention heads, and the layer-wise methods hypothesize that knowledge is represented in different layers of PLMs.

\paragraph{Gradient-based} ~\citet{dai2021knowledge} first introduce the concept of knowledge neurons, which are neurons in transformer~\citep{vaswani2017attention} related to certain factual knowledge. Specifically, they hypothesize the knowledge neurons are located in feed-forward networks, which are considered as key-value memories~\citep{geva2020transformer}. Then by feeding the LM with knowledge-expressing prompts such as ``Michael Jordan was born in [MASK]'', the corresponding knowledge neuron is identified as the neurons in the feed-forward networks with higher attribution scores, which are calculated based on integrated gradients.

\paragraph{Causal-inspired} 
~\citet{meng2022locating} identify knowledge neurons as the neuron activations in transformers that have the strongest causal effect on predicting certain factual knowledge. Such neurons are located through a causal mediation analysis. Specifically, they calculate the causal effect on factual prediction by comparing probability variation of object prediction between the clean and corrupted token embedding. Their experiments also demonstrate that the mid-layer feed-forward modules play a decisive role in factual knowledge representation.

\paragraph{Attention-based} 
In addition to the feed-forward layers, the attention heads are also be considered as representations which may encode the knowledge-related information.
~\citet{DBLP:conf/blackboxnlp/ClarkKLM19,htutAttentionHeadsBERT2019} investigate the linguistic knowledge encoded in attention heads, and find that while some individual attention heads are associated with specific aspects of syntax, the linguistic knowledge is distributed and represented by multiple attention heads.
~\citet{linOpenSesameGetting2019} find that PLMs' attention weights could encode syntactic properties such as subject-verb agreement and reflexive dependencies, and higher layers represent these syntactic properties more accurately.



\paragraph{Layer-wise} 
~\citet{linOpenSesameGetting2019} conduct a layer-wise probing for linguistic knowledge, which trains a specific classifier for each layer, and find that the lower layers encode the positional information of tokens, and higher layers encode more compositional information.
~\citet{liu-etal-2019-linguistic} analyze the layerwise transferability of PLMs on a wide range of tasks and find that the middle layers usually have better performance and transferability.
~\citet{wallat2021bertnesia} proposes to probe the captured factual knowledge with LAMA~\citep{petroniLanguageModelsKnowledge2019} of each layer in PLMs, and finds that a significant amount of knowledge is stored in the intermediate layers. 
~\citet{juneja2022finding} also conduct a layer-wised factual knowledge analysis based on knowledge neuron~\citep{dai2021knowledge}, and demonstrate that most relational knowledge (e.g., Paris is the capital of ``some nation''.) can be attributed to the middle layers, which would be refined into facts (e.g., Paris is the capital of France.) in the last few layers.

\subsection{Discussions and Future Works} 
The above studies reach some consensus about knowledge representation in PLMs, including: 1) Factual knowledge can be associated with feed-forward modules in middle or higher layers. 2) Linguistic knowledge is distributed and represented in multiple attention heads, while a single attention head can only associate with a specific aspect of linguistics. 3) The lower layers of PLMs often encode the coarse-grained and general information of knowledge, while the fine-grained and task-specific knowledge are mostly stored in higher layers.
These findings are valuable for us to understand knowledge representation in language models but are also limited to specific knowledge types or model architectures. 
Therefore the knowledge representation in PLMs is still an open problem which needs further exploration.

In the future, several directions of knowledge representation in PLMs may lie in the following:
1) Because knowledge representation is a long-standing concern in cognitive science, neuroscience, psychology, and artificial intelligence, it is helpful to borrow ideas from other related areas and design cognitively-inspired analysis methods. 
2) Current knowledge representation studies in PLMs mostly focus on a specific type of knowledge and often result in local and specific conclusions. 
It is important to comprehensively investigate different types of knowledge together,e.g., compare the differences and commonalities of knowledge representations of different knowledge types, pre-training tasks, or model architectures, and come up with more universal and insightful conclusions.

%% file: sections/probing.tex
\section{Knowledge Probing}
\label{sec:prob}

Knowledge probing aims to assess how well pre-trained language models entail different kinds of knowledge.
A comprehensive and accurate assessment of PLMs' knowledge can help us identify and understand language models' capabilities and deficiencies, allow a fair comparison between LMs with different architectures and pre-training tasks, guide the improvement of a specific model, and select suitable models for different real-world scenarios.
In this section, we will first introduce existing benchmarks for knowledge probing, then introduce the representative prompt-based and feature-based probing methods and analyze their corresponding limitations, and discuss future directions.

\subsection{Benchmarks for Knowledge Probing}

\begin{table*}[!htp]
\centering
\resizebox{\textwidth}{!}{
\begin{tabular}{@{}llll@{}}
\toprule
\textbf{Method}        & \textbf{Benchmarks}  & \textbf{Knowledge Type}        & \textbf{Formulation}                              \\ \hline
\multirow{9}{*}{Prompt-based}  & LM diagnostics~\citep{ettingerWhatBERTNot2020}       & linguistic            & text filling               \\
                               & BLiMP~\citep{warstadt-etal-2020-blimp-benchmark}                & linguistic            & sentence scores comparison            \\ \cdashline{2-4}
                               & LAMA~\citep{petroniLanguageModelsKnowledge2019}                 & factual, commonsense  & \multirow{4}{*}{text filling}       \\
                               & X-FACTR~\citep{jiangXFACTRMultilingualFactual2020}              & factual, multilingual &                                     \\
                               & Multilingual LAMA~\citep{kassner2021multilingual}    & factual, multilingual &                                     \\
                               & Bio LAMA~\citep{sung2021can}             & factual, biological   &                                     \\ \cdashline{2-4}
                               & CAT~\citep{Zhou2020EvaluatingCI}                  & commonsense           & sentence scores comparison            \\
                               & NumerSense~\citep{lin2020birds}   & commonsense, numerical & text filling \\
                               \cdashline{2-4}
                               & oLMPICS~\citep{Talmor2020oLMpicsOnWL}              & reasoning             & multiple choices            \\
                    
                               \hline
                               
\multirow{7}{*}{Feature-based} & Open Sesame~\citep{linOpenSesameGetting2019}          & linguistic            & diagnostic classifier and attention \\
                               & LKT~\citep{liuLinguisticKnowledgeTransferability2019}        & linguistic            & token or token pair labeling \\
                               & NPI probe~\citep{warstadtInvestigatingBERTKnowledge2019}            & linguistic            & probing classifier                  \\
                               & Edge probe~\citep{tenneyWhatYouLearn2019}           & linguistic, semantic  & edge probing                        \\
                               & MDL probe~\citep{voita2020information} & linguistic & minimum description length \\
                               \cdashline{2-4}
                               & Structural probe~\citep{hewittStructuralProbeFinding}    & syntactic                & structural probing                    \\
                               \cdashline{2-4}
                               & Physical Commonsense~\citep{forbesNeuralLanguageRepresentations2019} & commonsense, physical & probing classifier                   \\
\bottomrule
\end{tabular}
}
\caption{Summary about some representative knowledge probing benchmarks.}
\label{tab:probing}
\end{table*}

To assess the knowledge in PLMs, lots of benchmarks have been proposed to probe various knowledge contained in PLMs, for example, linguistic knowledge~\citep{ettingerWhatBERTNot2020,warstadt-etal-2020-blimp-benchmark,linOpenSesameGetting2019,warstadtInvestigatingBERTKnowledge2019,tenneyWhatYouLearn2019}, syntactic knowledge~\citep{DBLP:conf/blackboxnlp/ClarkKLM19,hewittStructuralProbeFinding}, factual knowledge~\citep{petroniLanguageModelsKnowledge2019,jiangXFACTRMultilingualFactual2020,kassner2021multilingual,sung2021can}, commonsense knowledge~\citep{forbesNeuralLanguageRepresentations2019,Zhou2020EvaluatingCI}, etc. Table~\ref{tab:probing} summarizes several representative knowledge probing benchmarks.






\subsection{Prompt-based Knowledge Probing}
Prompt-based probing is one of the most popular approaches for knowledge probing. To evaluate whether LMs know a specific knowledge such as the birthplace of Michael Jordan, we could query LMs with knowledge queries such as ``Michael Jordan was born in \underline{\hbox to 4mm{}}.'', where ``was born in'' is a prompt for a specific type of knowledge.
As shown in Table~\ref{tab:probing}, prompt-based probing has been widely used in benchmarks such as LAMA~\citep{petroniLanguageModelsKnowledge2019}, oLMpics~\citep{Talmor2020oLMpicsOnWL}, LM diagnostics~\citep{ettingerWhatBERTNot2020}, BIG-bench~\citep{srivastava2022beyond}, etc.

For prompt-based probing, the main challenge is how to design effective prompts which are suitable for different kinds of knowledge and different PLMs. 
In the following we will introduce the typical prompt types for knowledge probing and discuss their limitations.

\subsubsection{Prompt Development}

\paragraph{Handcraft Prompt} 
Early methods often manually write prompts for different kinds of knowledge. There are two primary advantages of manually created prompts: the readability and without the need of any other resources or training.
For example, LAMA~\citep{petroniLanguageModelsKnowledge2019} manually creates one cloze-style prompt for each relation, which is used to probe the factual knowledge in language models.
CAT~\citep{Zhou2020EvaluatingCI} reframe the instances in existing commonsense datasets into paired sentences with task-specific prompts, and determine whether PLMs contain specific commonsense knowledge by comparing the sentence scores, e.g., ``money can be used to buy cars'' v.s. ``money can be used to buy stars''.
oLMpics~\citep{Talmor2020oLMpicsOnWL} convert the probing tasks for reasoning ability into multi-choice questions with manually created prompts, and compare the LMs' probability of candidate choices.

\paragraph{Optimized Discrete Prompt}
Despite the mentioned advantages, ~\citet{jiangHowCanWe2020} argues that handcraft prompts could be sub-optimal. 
Therefore, a series of studies have been proposed to optimize the prompts in a discrete space so that PLMs could achieve better performance.
~\citet{jiangHowCanWe2020} propose a mining-based method in order to find prompts with higher performance from text corpus. They first retrieve potential prompts which contain both the subject and object entity, then select prompts using a validation dataset.
~\citet{feldmanCommonsenseKnowledgeMining2019} select prompt from a handcrafted candidate set according to the log-likelihood calculated by LMs.
~\citet{haviv-etal-2021-bertese} propose a paraphrasing-based method, where each query is first reframed by a trained rewriter and then fed into PLMs.
~\citet{shinAutoPromptElicitingKnowledge2020} propose an automatic prompt generation method based on gradient-guided search, where a prompt is iteratively updated from ``[MASK]'' token by maximizing the label likelihood of training instances. 





\paragraph{Continual Prompt}
Although the prompts generated by ~\citet{shinAutoPromptElicitingKnowledge2020} are discrete text, they are very difficult to be understood by humans. 
Therefore several studies directly search better-performed prompts on continual space rather than confining to discrete space, i.e., representing prompts as dense vectors. 
Continual prompts have shown good performance for knowledge probing, and further extensions including handcraft prompts initialization~\citep{zhong2021factual}, adding continual prompts on both input and transformer blocks~\citep{liPrefixTuningOptimizingContinuous2021} or adding LSTM layers above the input embeddings~\citep{liu2021gpt}.

\subsubsection{Limitations of Prompt-based Probing}

Although prompts have been widely used to probe the knowledge in PLMs, there are still lots of pending issues unresolved, which make the probing results unstable and the the assessment of knowledge in PLMs unreliable.

\paragraph{Inconsistent}
Prompt-based probing have been shown often result in inconsistent results due to prompt selection, instance verbalization, negation, etc. 
Firstly, ~\citet{elazar2021measuring} find semantically equivalent prompts may result in different predictions, ~\citet{cao-etal-2022-prompt} further find that PLMs would prefer specific prompts with the same linguistic regularity with the pre-training corpus, such a prompt preference will significantly affect the probing results, and result in inconsistent comparisons between PLMs.
Besides prompts, the instance verbalization process also leads to inconsistent predictions. For example, when we ask BERT ``The capital of the U.S. is [MASK]'', the answer is Washington, but when we replace the U.S. with its alias America, the prediction will change to Chicago.
In addition, PLMs also exhibit inconsistency when facing negation~\citep{kassnerNegatedMisprimedProbes2020b,jang2022can}. For instance, PLMs would generate highly similar predictions between a fact (``Birds can fly'') and its incorrect negation (``Birds cannot fly'')~\citep{kassnerNegatedMisprimedProbes2020b}. 
~\citet{jang2022can} conduct the negation experiments on PLMs of varying sizes and various downstream tasks, and find that not only PLMs cannot well understand negation prompts, but also show an inverse scaling law.

\paragraph{Inaccurate}
The performance of PLMs under prompt-based probing may also be overestimated. 
~\citet{poernerEBERTEfficientYetEffectiveEntity2020} find that many samples in the probing datasets could be easily ``guessed'' by only relying on the surface form association.
For example, the object entity is a substring of the subject entity (e.g., ``Apple Watch is produced by Apple'').
Furthermore, the training dataset for prompt optimization may correlate with probing dataset, which results in spurious correlations ~\citep{zhong2021factual} and the performance improvements may come from these spurious correlations.
~\citet{cao-etal-2021-knowledgeable} also find that many prompts with better performance are prompts which over-fit to answer distributions, rather than a better semantic description of the target relation.

\paragraph{Unreliable}
To reach a faithful probing results, it is essential to understand why PLMs make a specific prediction. However, studies find that PLMs do not always make predictions based on specific knowledge. In that case, the knowledge probing results could be unreliable.
~\citet{cao-etal-2021-knowledgeable} find that the prompts but not the answers dominate the prediction distribution of PLMs,  resulting in severely prompt-biased probing conclusions.
~\citet{li-etal-2022-pre} conduct a causal-inspired analysis and find that PLMs' predictions rely more on words that are close in position and frequently co-occur, rather than those related to knowledge.

\paragraph{Bias Analysis}
While lots of studies conduct empirical experiments on the biases in prompt-based probing, few have investigated the source and interpretation of these biases. 
Several studies employ causal analysis for bias analysis, which has been widely used to identify undesirable biases and fairness concerns~\citep{DBLP:conf/nips/HardtPNS16, DBLP:conf/nips/KilbertusRPHJS17,DBLP:conf/nips/KusnerLRS17,Vig2020InvestigatingGB,feder2021causal}.
~\citet{cao-etal-2022-prompt} propose a causal analysis framework to identify, interpret and eliminate biases that exist in prompt-based probing with a theoretical guarantee.
Similarly, ~\citet{elazar2022measuring} also propose a causal framework to estimate the causal effects of the data statistics in training corpus on the factual predictions of PLMs.
~\citet{finlayson2021causal} apply causal mediation analysis to investigate the syntactic agreement mechanisms in PLMs.

\subsection{Feature-based Knowledge Probing}

Feature-based knowledge probing is also widely used to probe the knowledge in PLMs, where the parameters of original PLMs are frozen, and the probing tasks are accomplished based on the internal representation or attention weights produced by PLMs. 
In this section, we introduce and discuss the feature-based probing approaches. 

\subsubsection{Classifier-based Probing}

Classifier-based probing trains a classifier to predict specific knowledge properties on the top of the fixed PLMs, and assesses the effectiveness of PLMs using the classifier's performance ~\citep{belinkov2022probing}. 
Such approaches are first propose to evaluate the linguistic properties (e.g., morphological, syntactic) associated with static embeddings~\citep{kohn-2015-whats,gupta-etal-2015-distributional}, and have been widely used to probe the linguistic knowledge~\citep{linOpenSesameGetting2019,tenneyWhatYouLearn2019,DBLP:conf/blackboxnlp/ClarkKLM19,liu-etal-2019-linguistic,hewittStructuralProbeFinding} and
 semantic knowledge~\citep{tenneyWhatYouLearn2019,wallaceNLPModelsKnow2019,yaghoobzadeh2019probing,liu-etal-2019-linguistic} in PLMs. 
Popular classifiers include linear classifier, logistics regression, multi-layer perceptron, etc.

\subsubsection{Classifier-free Probing}

Since the results and conclusions of classifier-based methods are dependent on the training quality and selection of the classifier,
some studies have developed feature-based probing approaches without an additional classifier.
For example, ~\citet{wu2020perturbed} propose perturbed masking, which calculates an impact matrix through a two-stage perturbation, where the matrix captures the impacts a token has on the prediction of another token, and is further used for the syntactic probe.
~\citet{zhou-srikumar-2021-directprobe} introduce DirectProbe, which directly probes the geometric properties of PLMs' representation without an additional classifier.
~\citet{DBLP:conf/blackboxnlp/ClarkKLM19} probe syntactic knowledge in language models by investigating the attention weights without a classifier, e.g., analyze the most attended word of the given token.

\subsubsection{Limitations of Feature-based Probing}
There are two main limitations of current feature-based probing approaches~\citep{rogers2020primer,belinkov2022probing}.
The first limitation concerns the attribution of results, which is originally pointed out by~\citet{hewittStructuralProbeFinding}. While most probes introduce additional training data and parameters, it's difficult to attribute evaluation results to the knowledge in PLMs, or the probe itself, which may learn to perform the probing task.
The second limitation pertains to the inconsistency between different probe designs for the same type of knowledge.  
There are various probe selections for each kind of knowledge,
but the probe results between simple probes like linear classifier or complex probes could be inconsistent.

\subsection{Discussions and Future Works}

With the growing scale and abilities of big language models, the comprehensive, accurate and reliable measurements of the actual knowledge and capabilities of LMs become increasingly important.
However, the accurate, robust and reliable probing approach is still an open problem.
Firstly, as we discussed above, both prompt-based probing and feature-based probing have their own limitations, which might result in unreliable or even contradicting conclusions.
Secondly, most existing benchmarks are specialized to specific knowledge types and specific model architectures.

In the future, the main directions of knowledge probing may lie in:
1) Comprehensive benchmark construction.
As we demonstrate in table~\ref{tab:probing}, current knowledge probing benchmarks are mostly too specialized, which may lead to inconsistent, biased or unreliable results. Therefore it is critical to build a comprehensive and unbiased benchmark.
2) Debiased probing approaches.
Currently prompt-based probing is the dominant knowledge probing methods due to its simplicity. However, there still exist lots of issues in prompt-based probing. Therefore, the design of unbiased datasets and better probing frameworks is another useful direction worth investigating.



%% file: sections/editing.tex
\section{Knowledge Editing}
\label{sec:edit}

Knowledge editing is the process which modifies the stored  knowledge in pre-trained language models, either by replacing it with new knowledge (e.g., changing the current prime minister of the UK to Rishi Sunak) or by removing it entirely (e.g., some personal privacy information).
There are two primary motivations for editing knowledge in language models: 1) even the state-of-the-art language models (e.g. ChatGPT\footnote{\url{https://openai.com/blog/chatgpt/}}) could learn lots of incorrect knowledge; 2) many facts are time-sensitive, requiring regular updates to their corresponding knowledge.

Unfortunately, editing knowledge in PLMs poses significant challenges. 
Firstly, naive solutions such as retraining are often impractical due to the massive size of large-scale language models. Secondly, due to the black box and non-linear nature of PLMs, any minor modification might result in a significant undesirable change in model predictions. As a result, it can be challenging to precisely edit the target knowledge.

To promote the development of relevant studies, ~\citet{de2021editing} formulate three desiderata for knowledge editing methods: 1) \textbf{Generality}: the method is able to edit the language models already pre-trained without the need for specialized re-training. 2) \textbf{Reliability}: the method is supposed to successfully edit knowledge required modification while not influencing the rest of knowledge in LMs. 3) \textbf{Consistency}: the modification should be consistent across paraphrases with equivalent semantics (e.g., Michael Jordan was born in [MASK]. v.s. The birthplace of Michael Jordan is [MASK].) and relevant knowledge required modification accordingly (e.g., Rishi Sunak becomes the prime minister of the UK. v.s. Liz Truss is not the primer minister of the UK.).

In this section, we divide current strategies for knowledge editing into four categories and the summary of comparisons between these approaches is shown in Table~\ref{tab:edit}. In the following we will describe and discuss these methods.


\begin{table*}[!tp]
\centering
\resizebox{\textwidth}{!}{
\begin{tabular}{lcccccc}
\toprule
\textbf{Approach}                & \textbf{\makecell{Knowledge \\ Support}} & \textbf{\makecell{Training\\Required}} & \textbf{\makecell{Online\\Edit}} & \textbf{\makecell{Batch\\Edit}} & \textbf{\makecell{Downstream\\Benifit}} & \textbf{\makecell{Unforeseen\\ Side Effects}} \\

\hline

\textbf{Constrained Tuning} &                            &                            &                      &                           &                             &                       \\
\quad FTM~\citep{zhu2020modifying}   & Factual       & YES &  NO  &   YES  & Potential &  YES\\ 

\cdashline{1-7}

\textbf{Memory-based}            &                            &                            &                      &                           &                             &                       \\
\quad SERAC~\citep{mitchell2022memory}   & Factual, QA &  YES & YES &   YES & NO    &  NO                       \\
\quad MEM-PROMPT~\citep{madaan2022memory}  & Linguistic, Ethics  &    NO  &  YES  &     YES  &    Potential                       &     Unlikely              \\
\quad CALINET~\citep{dong2022calibrating}   & Factual &       YES  &  NO &     YES  &     Potential & YES       \\

\cdashline{1-7}

\textbf{Meta-learning}           &                            &                            &                      &                           &                             &                       \\
\quad KNOWEDITOR~\citep{de2021editing}& Factual & YES & Possible  & YES  &       Potential  &   YES  \\
\quad MEMD~\citep{mitchell2021fast} &   Factual   &  YES    & NO  & YES  & Potential &  YES    \\

\cdashline{1-7}

\textbf{Locate and Edit}         &                            &                            &                      &                           &                             &                       \\
\quad Knowledge Neuron~\citep{dai2021knowledge}   &     Factual &  NO & YES & NO    &   NO & YES      \\
\quad ROME~\citep{meng2022locating}    & Factual &  NO & YES & NO    &   NO & Possible    \\ \bottomrule   
\end{tabular}
}
\caption{Comparisons between existing knowledge editing approaches. ``Online Edit'' refers to quickly editing an individual target knowledge. ``Batch Edit'' refers to editing a set of target knowledge simultaneously. ``Downstream Benefit'' refers to the potential for for the modified knowledge to be utilized by the edited language model for downstream tasks. ``Unforeseen Side Effects'' refers to the impact of knowledge editing on the language model beyond the modification of target knowledge.}
\label{tab:edit}
\end{table*}

\subsection{Constrained Fine-tuning}
The naive solution to edit knowledge in a PLM is to re-train it using the updated training dataset, but such a naive solution is computationally expensive and may be impractical because PLMs are involved.
Therefore, a better solution is to fine-tune PLMs only on a small subset which only contains the target samples. 
However, such a method may suffer from catastrophic forgetting, and affects the rest knowledge which is not intended to be edited.
Therefore, ~\citet{zhu2020modifying} propose to modify the knowledge in PLMs with constrained fine-tuning, specifically, 
they use an $\mathcal{L}_2$ or $\mathcal{L}_{\infty}$ normalization to constrain the parameters change of models.
Furthermore, they find that only fine-tuning the initial and final layers while keeping the rest of the model frozen yields better performance than fine-tuning the whole model.
However, in deep neural networks like PLMs, even a minor change of the parameters could change the model's predictions on a lot of samples. Therefore, such methods could potentially affect other knowledge stored in PLMs which is not required modification.

\subsection{Memory-Based Editing}
Instead of directly modifying parameters of PLMs, another natural solution is to maintain a knowledge cache which stores all new knowledge, and replace the original predictions when a input hits the cache. 
However, a symbolic knowledge cache may suffer from robustness issues, i.e., the inputs with the same meaning can differ in natural language expressions, therefore they may result in different predictions.
To address this problem, ~\citet{mitchell2022memory} propose a memory-based approach for knowledge editing. Specifically, the model contains five modules: an edit memory that stores the modified knowledge, a classifier, a counterfactual model, and the frozen original language model. Given an input, the classifier determines whether it hits a sample in the edit memory, and the counterfactual model's prediction will overrule the original language model's prediction if it hits a memory cache.
This method is effective but does not actually edit the knowledge encoded in the parameters of language models, thus cannot benefit downstream tasks.
Meanwhile, ~\citet{dong2022calibrating} add additional trainable parameters in the feed-forward module of PLMs, which are trained on a modified knowledge dataset while the original parameters are frozen. They also demonstrate that the modified knowledge could benefit related QA tasks.
Moreover, ~\citet{madaan2022memory} introduce the users' feedback for PLMs' error correction. Specifically, they maintain a memory of models' mistake and users' feedback, which enhance the model to produce updated prompt and avoid similar mistakes.


\subsection{Meta-learning-based Editing} 
~\citet{sinitsin2020editable} first propose editable training to conduct model editing based on meta-learning, which aims to train the model parameters to suit model editing. 
By constraining the training objective, the editing procedure could be accomplished under $k$ gradient step while ensuring reliability, locality, and efficiency.
However, such a method is not practical for pre-trained language models since it requires expensive specialized retraining.
A different strategy is to utilize a hyper network, which uses one network to generate the weights of another network~\citep{ha2016hypernetworks}. 
~\citet{de2021editing,hase2021language} train a hyper-network to predict the parameter changes for each data point, with the constraint of editing target knowledge without affecting others.
Although computationally efficient, ~\citet{mitchell2021fast} argues that this method fails to edit very large models, and proposes model editor networks with gradient decomposition (MEND). Specifically, by decomposing the gradient of standard fine-tuning into a low-rank form, they could train multiple MLPs to generate local model parameter changes, without damaging models' predictions on unrelated knowledge.
Experiments show that MEND can be applied to large pre-trained models for fast model editing.
One limitation of existing meta-learning-based methods is that their robustness and generalization are still questionable, as they ensure locality by constraining the parameter space change or the predictions on specific datasets.
In that case,
the knowledge that requires no modifications or the knowledge that is related to edited knowledge but not paraphrasing could also be incorrect.

\subsection{Locate and Edit}
Based on the assumption that "knowledge is locally stored in PLMs", the ``locate and edit'' strategy first locates the parameters corresponding to specific knowledge, and edit them by directly replacing with updated ones.
This approach is also introduced in Section~\ref{rep_method}. 
~\citet{dai2021knowledge} present a case study of factual knowledge editing in PLMs with corresponding knowledge neurons. By directly modifying the value of knowledge neurons, they achieve knowledge editing with a relatively low but nontrivial success rate.
Although the editing procedure is straightforward once the corresponding knowledge neuron is located, this method has not proved its effectiveness on large-scale editing or the effects of unrelated knowledge.
Similarly, ~\citet{meng2022locating} first connect the knowledge required modification with a key-value pair in one of the middle MLP layers, and modify the corresponding knowledge by directly updating the key-value pair.
Since these methods are based on the locality hypothesis of factual knowledge, which has not been widely confirmed yet, the changes in certain parameters may affect irrelevant knowledge and lead to unexpected results.

\subsection{Discussions and Future Works}

To utilize pre-trained language models as a sustainable knowledge resource, the precise, effective, reliable and consistent knowledge editing is essential.
However, as discussed above, all current editing methods have their own limitations. 
Therefore it is worthwhile to enhance current methods and develop new knowledge editing strategies.

In the future, several useful directions of knowledge editing may lie in:
1) \textbf{Broader range of target knowledge.} 
As shown in Table~\ref{tab:edit}, current studies mostly focus on the editing of factual knowledge, which is relatively easy to formalize and evaluate. In the future, researchers could explore the editing methods towards other kinds of knowledge, and develop universal approaches which can edit all kinds of knowledge in the same way. 
2) \textbf{Comprehensive evaluation.} 
Current most knowledge editing studies are evaluated using metrics such as editing success rate on target knowledge, predictions invariance rate on unrelated knowledge for assessing generality, and accuracy on paraphrases of target knowledge for assessing consistency. 
However, we find that these metrics are limited to comprehensively evaluate the knowledge editing capability of different approaches.
For instance, most evaluations only sample unrelated knowledge from the same distribution of target knowledge. However, the influence of a knowledge edit could be much broader, e.g., affecting the performance on downstream tasks or the knowledge from other distributions and categories. 
In addition, as mentioned in~\citet{mitchell2021fast}, most studies measure the consistency of samples generated through back translation, which ignores the knowledge affected by knowledge editing except the paraphrases, e.g., the country with the largest population would be affected by the population modification of the countries. 
Therefore, it is important to design comprehensive benchmark which can better assess the capabilities of editing strategies.
3) \textbf{More effective editing approaches.} 
Ideally, a knowledge editing approach should satisfy the desiderata of generality, reliability and consistency, and can handle large-scale and individual knowledge editing tasks with high efficiency.
To this end, we may borrow ideas from other fields, such as meta-learning, continual learning, and life-long learning. Furthermore, it is useful to connect knowledge editing studies with knowledge representation studies (\S~\ref{sec:repre}).

%% file: sections/application.tex
\section{Knowledge Application}
\label{sec:application}

Knowledge application studies how to effectively distill and leverage the knowledge in PLMs for other applications.
Specifically, we divide knowledge applications into two categories: language models as knowledge bases and language models for downstream tasks, and in following we describe them in detail.

\subsection{Language Models as Knowledge Bases}

\begin{table*}[tp]
    \centering
    \begin{tabular}{lll}
    \toprule
    \textbf{Perspectives} & \textbf{Structured KB}     & \textbf{LMs-as-KBs}                  \\ \hline
    \textbf{Construction} &                            &                                      \\
    \quad Ontology/Schema       & Pre-defined                 & \textbf{Open-ended} \emojismile                 \\
    \quad Process               & Pipline                     & \textbf{End-to-End} \emojismile                 \\
    \quad Human Effort          & Data annotation  & \textbf{Self-supervised} \emojismile \\ 
    \quad Expert Knowledge      & Common                      & \textbf{Not required} \emojismile \\ 
    \hline
    \textbf{Coverage}     &                            &                                      \\ 
    \quad Domain                & Constrained                & \textbf{Open} \emojismile                        \\
    \quad Amount                & Limited                     & Potential                            \\
    \quad Knowledge Fusing      & Complex                   & \textbf{Easy} \emojismile                        \\ \hline
    \textbf{Interaction}  &                            &                                      \\
    \quad Query                 & Structured                  & \textbf{Natural Language} \emojismile            \\
    \quad Prediction            & \textbf{Deterministic} \emojismile     & Probabilistic                        \\
    \quad Rejection             & \textbf{Yes} \emojismile               & Hard                           \\
    \quad Editing               & \textbf{Easy} \emojismile              & Limited                              \\ \hline
    \textbf{Reliability}  &                            &                                      \\ 
    \quad Ambiguity             & \textbf{Low} \emojismile               & High                                 \\
    \quad Correctness           & \textbf{Relatively High} \emojismile   & Questionable                         \\
    \quad Current Practicality  & \textbf{Extensive} \emojismile         & Limited yet                          \\
    \bottomrule
    \end{tabular}
    \caption{The comparisons between conventional structured knowledge bases and using language models as knowledge bases (LMs-as-KBs). Part of this table is inspired by  ~\citet{razniewski2021language}. The advantages are marked in bold.
    From the table, we can easily find that although LMs-as-KBs are more advantageous on construction and coverage, the critical current limitations of interaction and reliability significantly hinder its real-word applications, and far from substitution of structured knowledge bases.
    }
    \label{tab:kbs}
\end{table*}

The impressive performance of large-scale pre-trained language models, as well as the potentially enormous amount of implicitly stored knowledge, raises extensive attention about using language models as an alternative to conventional structured knowledge bases (LMs-as-KBs)~\citep{petroniLanguageModelsKnowledge2019,heinzerling2020language,jiangHowCanWe2020,wang2020language,cao-etal-2021-knowledgeable,razniewski2021language,alkhamissi2022review}. 

Unfortunately, along with the promising advantages and potentials compared with structured knowledge bases, there also exist intrinsic flaws for language models as knowledge base~\citep{razniewski2021language}, which are summarized in Table~\ref{tab:kbs}. In following we describe them in detail.

\textbf{Construction}  procedure is one of the biggest advantages of LMs-as-KBs compared with structured KBs.
Constructing large-scale structured KBs such as Freebase~\citep{10.1145/1376616.1376746} and Wikidata~\citep{10.1145/2629489} often requires extremely complex pipelines~\citep{petroniLanguageModelsKnowledge2019}, e.g., ontology construction, knowledge acquisition, knowledge verification, knowledge fusion, knowledge storage, and knowledge population. 
Such a complex pipeline involves lots of NLP techniques, including ontology engineering, entity linking, entity recognition, relation extraction, entity matching and so on. 
And each technique requires corresponding expert knowledge, supervised data and human efforts. 
Moreover, due to the pipeline nature, error propagation is always a critical issue.

In contrast, the knowledge of language models can be easily learned from pure text using self-supervised learning, without any explicit supervision signal (\S\ref{sssec:from_text}). Furthermore, the construction procedure is end-to-end, therefore no ontology engineering, expert knowledge, or human annotations are needed.

\textbf{Coverage} is another big advantage of LMs-as-KBs. Traditional structured KBs are often limited by its pre-defined schemas, and the difficulty of acquiring knowledge further limits their coverage. 
In comparison, by directly representing knowledge in parameters, there is no schema limitations for LMs-as-KBs. And all knowledge is learned from un-annotated text corpus, therefore the knowledge coverage is mostly only determined by the coverage of pre-training corpus.

The above advantages make LMs-as-KBs an extremely attractive and promising idea.
However, there are also some intrinsic flaws which hinder LMs from fully substituting structured KBs.

\textbf{Interaction} with structured KB and LMs-as-KBs are quite different.
Structured KBs often use structural querying methods such as SPARQL~\citep{perez2009semantics}, e.g., querying the birthplace of Michael Jordan using $<$Michael Jordan, Birthplace, ?$>$.
In the case of language model-based KBs, the queries are mostly natural language expressions such as ``The birthplace of Michael Jordan is [MASK]''.

Compared with structural queries, natural language-based queries are more natural and friendly for users. 
However, structured KBs can return deterministic answers (e.g., Brooklyn), but LM-based KBs can only generate candidates with different probabilities (e.g., <Brooklyn, 0.8>). The probabilistic predictions may be incorrect, inconsistent and confusing. 
Furthermore, structured KBs can identify the queries they cannot answer, but current LM-based KBs can hardly reject the queries it cannot answer, thus resulting in the knowledge hallucination problem.
Concretely, if we query some knowledge that is not stored in a structured KB, the answer could be blank when no tuples are matched. However, no matter what we ask, language models will always ``guess'' the answers, even such knowledge is never learned by LMs. 
Although there are some naive solutions to this problem such as rejecting answers with a low probability, this is still a open problem currently.

Finally, it is difficult to edit knowledge in LM-based KBs, as discussed in Section~\ref{sec:edit}. In comparison, it is easy to add, modify and delete knowledge in structured KBs.

\textbf{Reliability} is another concern for LMs-as-KBs. 
The first problem is ambiguity. In structured KBs, all entities and facts have their own IDs (e.g., Q89 for Apple the fruit and Q312 for Apple Inc. in Wikidata), therefore there is no ambiguity problem. 
However, in LM-based KBs, all pieces of knowledge are represented as natural language expressions and will therefore suffer from the ambiguity problem of natural language. 
For example, do "U.S.A" and "America" represent the same entity in a language model?
Previous studies have observed that such verbalization requirements will result in prompt preference bias and instance verbalization bias in LMs-as-KBs ~\citep{cao-etal-2022-prompt}.
The consistency of predictions is another drawback of LMs-as-KBs, i.e., a LM-based KB may return different answers to the semantically equivalent queries.

\subsection{Language Models for Downstream Tasks}



\begin{figure*}[!htp]
    \centering
    \includegraphics[width=\textwidth]{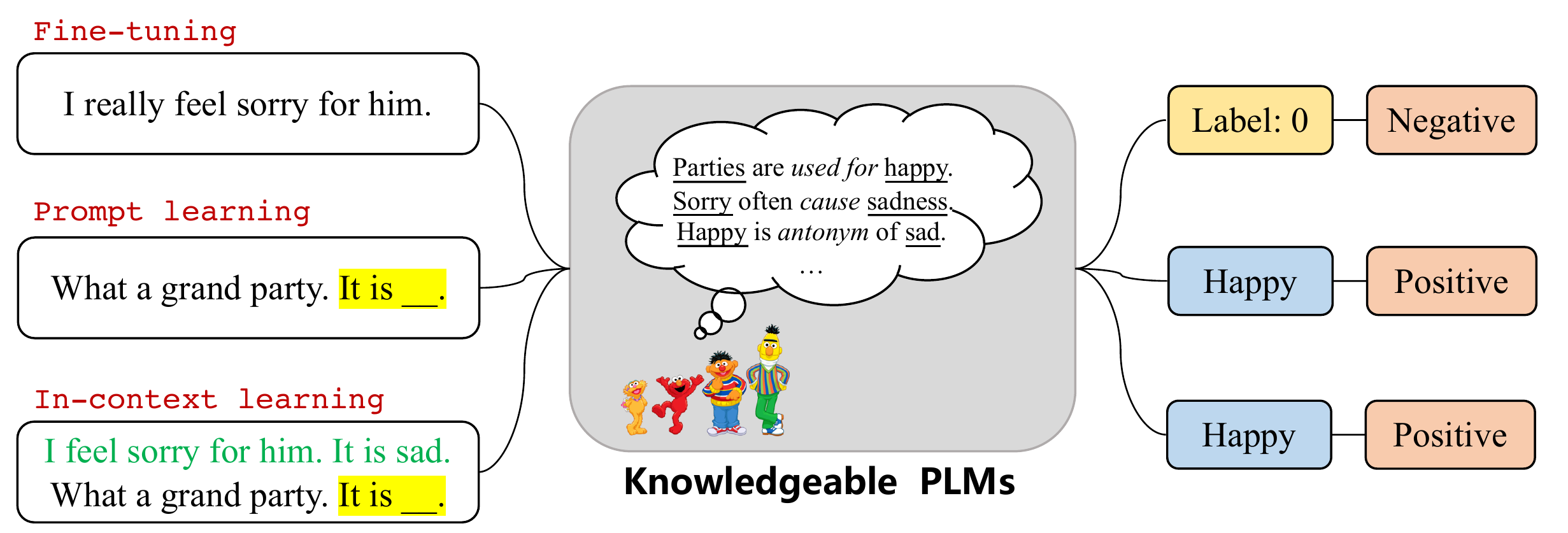}
    \caption{The primary paradigms that apply the knowledge in PLMs to downstream tasks.}
    \label{fig:app_3}
\end{figure*}

Besides using language models as knowledge bases, the knowledge in PLMs can also benefit many downstream tasks in different ways. Fig.~\ref{fig:app_3} shows three main paradigms and we describe them in detail.

\subsubsection{Fine-tuning}

Fine-tuning is a common way to leverage knowledge in language models, which learns to distill and leverage knowledge by further tuning PLMs using task-specific datasets.
Firstly, implicitly learned knowledge from text has been recognized as one of the main reasons for PLMs' remarkable performance and strong generalization ability across so many NLP tasks~\citep{manning2020emergent,wei2021knowledge,yang2021survey,yin2022survey}.
Secondly, many studies have shown that injecting knowledge into language models can lead to better performance on downstream tasks.
For instance, integrating entity knowledge into PLMs could improve the performance of a wide range of language understanding tasks~\citet{sun2019ernie, shen2020exploiting}, infusing factual knowledge into PLMs could benifit their performance on tasks such as relation extraction, entity typing, etc.~\citep{zhang2019ernie, wang2021kepler, wang2020k, liu2020k}, and incorporating linguistic knowledge with PLMs could increase their performance on benchmarks such as GLUE~\citep{levine2019sensebert,sachan2020syntax,bai2021syntax}.

\subsubsection{Prompt Learning}
Prompt-based learning is another way to leverage the knowledge in PLMs for downstream tasks.
For example, to classify the sentiment polarity of the sentence ``Best movie ever.'', we can add a prompt and transform the input into ``Best movie ever. It is \underline{\hbox to 4mm{}}.''. 
And the polarity can be determined by comparing the PLMs' prediction probability between candidate answers ``good'' and ``bad''.
By selecting appropriate prompts, PLMs have been shown competitive zero-shot performance on some downstream tasks without any supervised training~\citep{radfordLanguageModelsAre2019,brown2020language,liu2021pre}. 

Because handcraft prompts often suffer from unstable performance across different prompts and cannot utilize the information from supervised data, many prompt optimization approaches have been proposed to acquire better-performing prompts~\citep{liu2021pre}, such as paraphrasing~\citep{jiangHowCanWe2020, haviv-etal-2021-bertese}, gradient-based search~\citep{shinAutoPromptElicitingKnowledge2020}, model generation~\citep{gao2020making}, knowledge enhanced~\citep{hu2021knowledgeable}, etc.
Furthermore, prompt-tuning, which adds some trainable vectors to the inputs as continuous prompts, while keeping the parameters of LMs freezing, has achieved competitive performance with fine-tuning~\citep{li2021prefix,liu2021gpt,hambardzumyan2021warp,lester2021power}.
In addition to optimizing single prompts, ensembling~\citep{jiangHowCanWe2020,qin2021learning}, compositing~\citep{han2021ptr}, or decoupling~\citep{ozturkler2022thinksum} multiple prompts could also improve model performance. 
Moreover, prompt has also been applied to data augmentation~\citep{schick2020exploiting}, domain adaptation~\citep{ben2021pada}, debiasing~\citep{schick2021self} and so on.

More recently, instruction-tuning, which pre-trains LMs on a wide range of datasets given the natural language description of tasks as instructions, has achieved significant performance and generalization ability improvements of language models~\citep{wei2021finetuned,sanh2021multitask,ouyang2022training,chung2022scaling}.

\subsubsection{In-context Learning}

\paragraph{Applications} 
Currently the parameters of PLMs have been scaled to 175B (e.g., GPT-3~\citep{brown2020language}, OPT~\citep{zhang2022opt}, BLOOM\footnote{\url{https://huggingface.co/bigscience/bloom}}) or even larger (e.g., PaLM~\citep{chowdhery2022palm}), making
the computational expense of fine-tuning and prompt-tuning infeasible for most researchers. 
Therefore, tuning-free in-context learning has become one of the most popular approaches to apply the knowledge in large-scale PLMs in downstream tasks~\citep{dong2022survey}.
For instance, for the sentiment classification task, in-context learning will first sample several demonstrations, such as (what a horrible meal, negative), and combine them with the original query. In this way, the input becomes ``What a horrible meal. It is bad. [SEP] Best movie ever. It is \underline{\hbox to 4mm{}}.''
The provided demonstrations offer extra information about the task and enable PLMs to utilize the analogy ability to predict the correct answer.
In-context learning has achieved good performance on lots of downstream tasks such as language understanding~\citep{brown2020language,zhao2021calibrate,lee2021good,eisenstein2022honest,zhang2022robustness}, data generation~\citep{li2022explanations,dai2022promptagator,yu2022generate}, or reasoning~\citep{wei2022chain,lampinen2022can,zhou2022least}.

\paragraph{Bias Problem}
One drawback of in-context learning is the bias problem, i.e.,
the performance is sensitive to demonstration selections, demonstration orders, label distribution of demonstrations and prompt selection, etc.~\citep{zhao2021calibrate,lu2021fantastically,liu2021makes}.
Therefore, to achieve better performance of in-context learning,
~\citet{zhao2021calibrate} first propose to estimate the biases by feeding the model with an uninformative input (e.g., [MASK] or N/A), and then calibrate the prediction probabilities uniformly distributed for eliminating the models' bias towards specific answers.
For demonstration selection, ~\citet{gao2020making,liu2021makes} propose to select demonstrations that are semantically close to the input query.
~\citet{rubin2021learning} train a dense retriever on LM-scored datasets to select demonstrations.
~\citet{su2022selective} introduce a graph-based selection method to ensure the demonstration's diversity and representativeness.
For demonstration sort, ~\citet{lu2021fantastically} first construct a development dataset by sampling from language models, and then use entropy-based metrics to determine the optimal demonstration permutation.
For prompt selection, ~\citet{gao2020making} use a language model to generate candidate prompts and select ones with better performance on the development set.






\paragraph{Mechanism}
Although in-context learning has been widely applied on various downstream tasks, its underlying mechanism is still unclear.
~\citet{reynolds2021prompt} find that zero-shot prompting sometimes can significantly outperform in-context learning, and argue that the additional demonstrations do not help PLMs to learn a new task, but rather locate the task they have already learned.
~\citet{cao-etal-2021-knowledgeable} investigate the in-context learning for knowledge probing, and find that the demonstrations can only provide type-level guidance but factual information.
~\citet{min2022rethinking} find that randomly replacing the demonstrations' labels hardly affects the performance, and show that the effectiveness of in-context learning relies more on the label space and input distribution restriction provided by demonstrations rather than the precise input-label mapping. 
~\citet{DBLP:journals/corr/abs-2205-05055} find that only when the data includes both burstiness and large-scale of rarely occurring classes, in-context learning capability can emerge in transformer model.
\citet{von2022transformers} investigate the connections between in-context learning and gradient descent, and demonstrate the similarity between in-context learning and the gradient-based few-shot learning.








\subsection{Discussions and Future Works}

Leveraging knowledge in PLMs is both promising and challenging.
On the one hand, it is obvious that the large amount of implicit knowledge stored in PLMs will benefit different downstream tasks.
On the other hand, all current application paradigms have their own limitations. 
For instance, the consistency and reliability of LMs-as-KBs hinder PLMs to replace structured KBs, 
Moreover, fine-tuning, prompt learning and in-context learning methods often suffer from catastrophic forgetting, computational cost, inconsistent and unstable predictions, social bias, etc.

To address these challenges, several main future directions of knowledge application may lie in the following:
1) For LMs-as-KBs, we need to propose specific pre-training approaches to address current shortcomings in consistency and reliability.
2) For LMs for downstream tasks, we suggest explore more application strategies, such as new tuning-free methods to address the computational cost issue and black-box tuning~\citep{sun2022black} methods to tune pre-trained language models without access to their parameters.

%% file: sections/conclusion.tex
\section{Conclusions}
\label{sec:conc}

In this survey, we conduct a comprehensive review about the life circle of knowledge in pre-trained language models, including knowledge acquisition, knowledge representation, knowledge probing, knowledge editing and knowledge application.
We systematically review related studies for each period, discuss the advantages and limitations of different methods, summarize the main challenge, and present some future directions. We believe this survey will benefit researchers in many areas such as language models, knowledge graph, knowledge base, etc.

\section*{Acknowledgments}
We sincerely thank all anonymous reviewers for their insightful comments and valuable suggestions. This research work is supported by the National Natural Science Foundation of China under Grants no. 62122077 and CAS Project for Young Scientists in Basic Research under Grant No.YSBR-040.

%% file: acl_latex.bbl
\begin{thebibliography}{177}
\expandafter\ifx\csname natexlab\endcsname\relax\def\natexlab#1{#1}\fi

\bibitem[{Achille et~al.(2019)Achille, Rovere, and
  Soatto}]{achille2018critical}
Alessandro Achille, Matteo Rovere, and Stefano Soatto. 2019.
\newblock \href {https://openreview.net/forum?id=BkeStsCcKQ} {Critical learning
  periods in deep networks}.
\newblock In \emph{Proc. of ICLR}. OpenReview.net.

\bibitem[{AlKhamissi et~al.(2022)AlKhamissi, Li, Celikyilmaz, Diab, and
  Ghazvininejad}]{alkhamissi2022review}
Badr AlKhamissi, Millicent Li, Asli Celikyilmaz, Mona Diab, and Marjan
  Ghazvininejad. 2022.
\newblock \href {https://arxiv.org/abs/2204.06031} {A review on language models
  as knowledge bases}.
\newblock \emph{ArXiv preprint}, abs/2204.06031.

\bibitem[{Bai et~al.(2021)Bai, Wang, Chen, Yang, Bai, Yu, and
  Tong}]{bai2021syntax}
Jiangang Bai, Yujing Wang, Yiren Chen, Yaming Yang, Jing Bai, Jing Yu, and
  Yunhai Tong. 2021.
\newblock \href {https://doi.org/10.18653/v1/2021.eacl-main.262}
  {Syntax-{BERT}: Improving pre-trained transformers with syntax trees}.
\newblock In \emph{Proceedings of the 16th Conference of the European Chapter
  of the Association for Computational Linguistics: Main Volume}, pages
  3011--3020, Online. Association for Computational Linguistics.

\bibitem[{Baldini~Soares et~al.(2019)Baldini~Soares, FitzGerald, Ling, and
  Kwiatkowski}]{baldini-soares-etal-2019-matching}
Livio Baldini~Soares, Nicholas FitzGerald, Jeffrey Ling, and Tom Kwiatkowski.
  2019.
\newblock \href {https://doi.org/10.18653/v1/P19-1279} {Matching the blanks:
  Distributional similarity for relation learning}.
\newblock In \emph{Proc. of ACL}, pages 2895--2905, Florence, Italy.
  Association for Computational Linguistics.

\bibitem[{Banerjee and Baral(2020)}]{banerjee2020self}
Pratyay Banerjee and Chitta Baral. 2020.
\newblock \href {https://doi.org/10.18653/v1/2020.emnlp-main.11}
  {Self-supervised knowledge triplet learning for zero-shot question
  answering}.
\newblock In \emph{Proc. of EMNLP}, pages 151--162, Online. Association for
  Computational Linguistics.

\bibitem[{Belinkov(2022)}]{belinkov2022probing}
Yonatan Belinkov. 2022.
\newblock \href {https://doi.org/10.1162/coli_a_00422} {Probing classifiers:
  Promises, shortcomings, and advances}.
\newblock \emph{Computational Linguistics}, 48(1):207--219.

\bibitem[{Ben-David et~al.(2021)Ben-David, Oved, and Reichart}]{ben2021pada}
Eyal Ben-David, Nadav Oved, and Roi Reichart. 2021.
\newblock \href {https://arxiv.org/abs/2102.12206} {Pada: A prompt-based
  autoregressive approach for adaptation to unseen domains}.
\newblock \emph{ArXiv preprint}, abs/2102.12206.

\bibitem[{Bollacker et~al.(2008)Bollacker, Evans, Paritosh, Sturge, and
  Taylor}]{10.1145/1376616.1376746}
Kurt Bollacker, Colin Evans, Praveen Paritosh, Tim Sturge, and Jamie Taylor.
  2008.
\newblock \href {https://doi.org/10.1145/1376616.1376746} {Freebase: A
  collaboratively created graph database for structuring human knowledge}.
\newblock In \emph{Proceedings of the 2008 ACM SIGMOD International Conference
  on Management of Data}, SIGMOD '08, page 1247–1250, New York, NY, USA.
  Association for Computing Machinery.

\bibitem[{Bosselut et~al.(2019)Bosselut, Rashkin, Sap, Malaviya, Celikyilmaz,
  and Choi}]{bosselut2019comet}
Antoine Bosselut, Hannah Rashkin, Maarten Sap, Chaitanya Malaviya, Asli
  Celikyilmaz, and Yejin Choi. 2019.
\newblock \href {https://doi.org/10.18653/v1/P19-1470} {{COMET}: Commonsense
  transformers for automatic knowledge graph construction}.
\newblock In \emph{Proc. of ACL}, pages 4762--4779, Florence, Italy.
  Association for Computational Linguistics.

\bibitem[{Bouraoui et~al.(2020)Bouraoui, Camacho{-}Collados, and
  Schockaert}]{bouraouiInducingRelationalKnowledge2019}
Zied Bouraoui, Jos{\'{e}} Camacho{-}Collados, and Steven Schockaert. 2020.
\newblock \href {https://aaai.org/ojs/index.php/AAAI/article/view/6242}
  {Inducing relational knowledge from {BERT}}.
\newblock In \emph{The Thirty-Fourth {AAAI} Conference on Artificial
  Intelligence, {AAAI} 2020, The Thirty-Second Innovative Applications of
  Artificial Intelligence Conference, {IAAI} 2020, The Tenth {AAAI} Symposium
  on Educational Advances in Artificial Intelligence, {EAAI} 2020, New York,
  NY, USA, February 7-12, 2020}, pages 7456--7463. {AAAI} Press.

\bibitem[{Brown et~al.(2020)Brown, Mann, Ryder, Subbiah, Kaplan, Dhariwal,
  Neelakantan, Shyam, Sastry, Askell, Agarwal, Herbert{-}Voss, Krueger,
  Henighan, Child, Ramesh, Ziegler, Wu, Winter, Hesse, Chen, Sigler, Litwin,
  Gray, Chess, Clark, Berner, McCandlish, Radford, Sutskever, and
  Amodei}]{brown2020language}
Tom~B. Brown, Benjamin Mann, Nick Ryder, Melanie Subbiah, Jared Kaplan,
  Prafulla Dhariwal, Arvind Neelakantan, Pranav Shyam, Girish Sastry, Amanda
  Askell, Sandhini Agarwal, Ariel Herbert{-}Voss, Gretchen Krueger, Tom
  Henighan, Rewon Child, Aditya Ramesh, Daniel~M. Ziegler, Jeffrey Wu, Clemens
  Winter, Christopher Hesse, Mark Chen, Eric Sigler, Mateusz Litwin, Scott
  Gray, Benjamin Chess, Jack Clark, Christopher Berner, Sam McCandlish, Alec
  Radford, Ilya Sutskever, and Dario Amodei. 2020.
\newblock \href
  {https://proceedings.neurips.cc/paper/2020/hash/1457c0d6bfcb4967418bfb8ac142f64a-Abstract.html}
  {Language models are few-shot learners}.
\newblock In \emph{Advances in Neural Information Processing Systems 33: Annual
  Conference on Neural Information Processing Systems 2020, NeurIPS 2020,
  December 6-12, 2020, virtual}.

\bibitem[{Cao et~al.(2022)Cao, Lin, Han, Liu, and Sun}]{cao-etal-2022-prompt}
Boxi Cao, Hongyu Lin, Xianpei Han, Fangchao Liu, and Le~Sun. 2022.
\newblock \href {https://doi.org/10.18653/v1/2022.acl-long.398} {Can prompt
  probe pretrained language models? understanding the invisible risks from a
  causal view}.
\newblock In \emph{Proc. of ACL}, pages 5796--5808, Dublin, Ireland.
  Association for Computational Linguistics.

\bibitem[{Cao et~al.(2021)Cao, Lin, Han, Sun, Yan, Liao, Xue, and
  Xu}]{cao-etal-2021-knowledgeable}
Boxi Cao, Hongyu Lin, Xianpei Han, Le~Sun, Lingyong Yan, Meng Liao, Tong Xue,
  and Jin Xu. 2021.
\newblock \href {https://doi.org/10.18653/v1/2021.acl-long.146} {Knowledgeable
  or educated guess? revisiting language models as knowledge bases}.
\newblock In \emph{Proc. of ACL}, pages 1860--1874, Online. Association for
  Computational Linguistics.

\bibitem[{Chan et~al.(2022)Chan, Santoro, Lampinen, Wang, Singh, Richemond,
  McClelland, and Hill}]{DBLP:journals/corr/abs-2205-05055}
Stephanie C.~Y. Chan, Adam Santoro, Andrew~K. Lampinen, Jane~X. Wang, Aaditya
  Singh, Pierre~H. Richemond, Jay McClelland, and Felix Hill. 2022.
\newblock \href {https://doi.org/10.48550/arXiv.2205.05055} {Data
  distributional properties drive emergent in-context learning in
  transformers}.
\newblock \emph{CoRR}, abs/2205.05055.

\bibitem[{Chiang et~al.(2020)Chiang, Huang, and Lee}]{chiang2020pretrained}
Cheng-Han Chiang, Sung-Feng Huang, and Hung-yi Lee. 2020.
\newblock \href {https://doi.org/10.18653/v1/2020.emnlp-main.553} {{P}retrained
  language model embryology: {T}he birth of {ALBERT}}.
\newblock In \emph{Proc. of EMNLP}, pages 6813--6828, Online. Association for
  Computational Linguistics.

\bibitem[{Chowdhery et~al.(2022)Chowdhery, Narang, Devlin, Bosma, Mishra,
  Roberts, Barham, Chung, Sutton, Gehrmann et~al.}]{chowdhery2022palm}
Aakanksha Chowdhery, Sharan Narang, Jacob Devlin, Maarten Bosma, Gaurav Mishra,
  Adam Roberts, Paul Barham, Hyung~Won Chung, Charles Sutton, Sebastian
  Gehrmann, et~al. 2022.
\newblock \href {https://arxiv.org/abs/2204.02311} {Palm: Scaling language
  modeling with pathways}.
\newblock \emph{ArXiv preprint}, abs/2204.02311.

\bibitem[{Chung et~al.(2022)Chung, Hou, Longpre, Zoph, Tay, Fedus, Li, Wang,
  Dehghani, Brahma et~al.}]{chung2022scaling}
Hyung~Won Chung, Le~Hou, Shayne Longpre, Barret Zoph, Yi~Tay, William Fedus,
  Eric Li, Xuezhi Wang, Mostafa Dehghani, Siddhartha Brahma, et~al. 2022.
\newblock \href {https://arxiv.org/abs/2210.11416} {Scaling
  instruction-finetuned language models}.
\newblock \emph{ArXiv preprint}, abs/2210.11416.

\bibitem[{Churchland and Sejnowski(1988)}]{churchland1988perspectives}
Patricia~S Churchland and Terrence~J Sejnowski. 1988.
\newblock Perspectives on cognitive neuroscience.
\newblock \emph{Science}, 242(4879):741--745.

\bibitem[{Clark et~al.(2019)Clark, Khandelwal, Levy, and
  Manning}]{DBLP:conf/blackboxnlp/ClarkKLM19}
Kevin Clark, Urvashi Khandelwal, Omer Levy, and Christopher~D. Manning. 2019.
\newblock \href {https://doi.org/10.18653/v1/W19-4828} {What does {BERT} look
  at? an analysis of {BERT}{'}s attention}.
\newblock In \emph{Proceedings of the 2019 ACL Workshop BlackboxNLP: Analyzing
  and Interpreting Neural Networks for NLP}, pages 276--286, Florence, Italy.
  Association for Computational Linguistics.

\bibitem[{Dai et~al.(2022{\natexlab{a}})Dai, Dong, Hao, Sui, Chang, and
  Wei}]{dai2021knowledge}
Damai Dai, Li~Dong, Yaru Hao, Zhifang Sui, Baobao Chang, and Furu Wei.
  2022{\natexlab{a}}.
\newblock \href {https://doi.org/10.18653/v1/2022.acl-long.581} {Knowledge
  neurons in pretrained transformers}.
\newblock In \emph{Proc. of ACL}, pages 8493--8502, Dublin, Ireland.
  Association for Computational Linguistics.

\bibitem[{Dai et~al.(2022{\natexlab{b}})Dai, Zhao, Ma, Luan, Ni, Lu, Bakalov,
  Guu, Hall, and Chang}]{dai2022promptagator}
Zhuyun Dai, Vincent~Y Zhao, Ji~Ma, Yi~Luan, Jianmo Ni, Jing Lu, Anton Bakalov,
  Kelvin Guu, Keith~B Hall, and Ming-Wei Chang. 2022{\natexlab{b}}.
\newblock \href {https://arxiv.org/abs/2209.11755} {Promptagator: Few-shot
  dense retrieval from 8 examples}.
\newblock \emph{ArXiv preprint}, abs/2209.11755.

\bibitem[{Davison et~al.(2019)Davison, Feldman, and
  Rush}]{feldmanCommonsenseKnowledgeMining2019}
Joe Davison, Joshua Feldman, and Alexander Rush. 2019.
\newblock \href {https://doi.org/10.18653/v1/D19-1109} {Commonsense knowledge
  mining from pretrained models}.
\newblock In \emph{Proc. of EMNLP}, pages 1173--1178, Hong Kong, China.
  Association for Computational Linguistics.

\bibitem[{De~Cao et~al.(2021)De~Cao, Aziz, and Titov}]{de2021editing}
Nicola De~Cao, Wilker Aziz, and Ivan Titov. 2021.
\newblock \href {https://doi.org/10.18653/v1/2021.emnlp-main.522} {Editing
  factual knowledge in language models}.
\newblock In \emph{Proc. of EMNLP}, pages 6491--6506, Online and Punta Cana,
  Dominican Republic. Association for Computational Linguistics.

\bibitem[{Devlin et~al.(2019)Devlin, Chang, Lee, and
  Toutanova}]{devlin2018bert}
Jacob Devlin, Ming-Wei Chang, Kenton Lee, and Kristina Toutanova. 2019.
\newblock \href {https://doi.org/10.18653/v1/N19-1423} {{BERT}: Pre-training of
  deep bidirectional transformers for language understanding}.
\newblock In \emph{Proc. of NAACL-HLT}, pages 4171--4186, Minneapolis,
  Minnesota. Association for Computational Linguistics.

\bibitem[{Dong et~al.(2022)Dong, Dai, Song, Xu, Sui, and
  Li}]{dong2022calibrating}
Qingxiu Dong, Damai Dai, Yifan Song, Jingjing Xu, Zhifang Sui, and Lei Li.
  2022.
\newblock \href {https://arxiv.org/abs/2210.03329} {Calibrating factual
  knowledge in pretrained language models}.
\newblock \emph{ArXiv preprint}, abs/2210.03329.

\bibitem[{Dong et~al.(2023)Dong, Li, Dai, Zheng, Wu, Chang, Sun, Xu, and
  Sui}]{dong2022survey}
Qingxiu Dong, Lei Li, Damai Dai, Ce~Zheng, Zhiyong Wu, Baobao Chang, Xu~Sun,
  Jingjing Xu, and Zhifang Sui. 2023.
\newblock \href {https://arxiv.org/abs/2301.00234} {A survey for in-context
  learning}.
\newblock \emph{ArXiv preprint}, abs/2301.00234.

\bibitem[{Eisenstein et~al.(2022)Eisenstein, Andor, Bohnet, Collins, and
  Mimno}]{eisenstein2022honest}
Jacob Eisenstein, Daniel Andor, Bernd Bohnet, Michael Collins, and David Mimno.
  2022.
\newblock \href {https://arxiv.org/abs/2210.02498} {Honest students from
  untrusted teachers: Learning an interpretable question-answering pipeline
  from a pretrained language model}.
\newblock \emph{ArXiv preprint}, abs/2210.02498.

\bibitem[{Elazar et~al.(2022)Elazar, Kassner, Ravfogel, Feder, Ravichander,
  Mosbach, Belinkov, Sch{\"u}tze, and Goldberg}]{elazar2022measuring}
Yanai Elazar, Nora Kassner, Shauli Ravfogel, Amir Feder, Abhilasha Ravichander,
  Marius Mosbach, Yonatan Belinkov, Hinrich Sch{\"u}tze, and Yoav Goldberg.
  2022.
\newblock \href {https://arxiv.org/abs/2207.14251} {Measuring causal effects of
  data statistics on language model'sfactual'predictions}.
\newblock \emph{ArXiv preprint}, abs/2207.14251.

\bibitem[{Elazar et~al.(2021)Elazar, Kassner, Ravfogel, Ravichander, Hovy,
  Sch{\"u}tze, and Goldberg}]{elazar2021measuring}
Yanai Elazar, Nora Kassner, Shauli Ravfogel, Abhilasha Ravichander, Eduard
  Hovy, Hinrich Sch{\"u}tze, and Yoav Goldberg. 2021.
\newblock \href {https://doi.org/10.1162/tacl_a_00410} {Measuring and improving
  consistency in pretrained language models}.
\newblock \emph{Transactions of the Association for Computational Linguistics},
  9:1012--1031.

\bibitem[{Ettinger(2020)}]{ettingerWhatBERTNot2020}
Allyson Ettinger. 2020.
\newblock \href {https://doi.org/10.1162/tacl_a_00298} {What {BERT} is not:
  Lessons from a new suite of psycholinguistic diagnostics for language
  models}.
\newblock \emph{Transactions of the Association for Computational Linguistics},
  8:34--48.

\bibitem[{Feder et~al.(2021)Feder, Keith, Manzoor, Pryzant, Sridhar,
  Wood-Doughty, Eisenstein, Grimmer, Reichart, Roberts
  et~al.}]{feder2021causal}
Amir Feder, Katherine~A Keith, Emaad Manzoor, Reid Pryzant, Dhanya Sridhar,
  Zach Wood-Doughty, Jacob Eisenstein, Justin Grimmer, Roi Reichart, Margaret~E
  Roberts, et~al. 2021.
\newblock \href {https://arxiv.org/abs/2109.00725} {Causal inference in natural
  language processing: Estimation, prediction, interpretation and beyond}.
\newblock \emph{ArXiv preprint}, abs/2109.00725.

\bibitem[{F{\'e}vry et~al.(2020)F{\'e}vry, Baldini~Soares, FitzGerald, Choi,
  and Kwiatkowski}]{fevry2020entities}
Thibault F{\'e}vry, Livio Baldini~Soares, Nicholas FitzGerald, Eunsol Choi, and
  Tom Kwiatkowski. 2020.
\newblock \href {https://doi.org/10.18653/v1/2020.emnlp-main.400} {Entities as
  experts: Sparse memory access with entity supervision}.
\newblock In \emph{Proc. of EMNLP}, pages 4937--4951, Online. Association for
  Computational Linguistics.

\bibitem[{Finlayson et~al.(2021)Finlayson, Mueller, Gehrmann, Shieber, Linzen,
  and Belinkov}]{finlayson2021causal}
Matthew Finlayson, Aaron Mueller, Sebastian Gehrmann, Stuart Shieber, Tal
  Linzen, and Yonatan Belinkov. 2021.
\newblock \href {https://doi.org/10.18653/v1/2021.acl-long.144} {Causal
  analysis of syntactic agreement mechanisms in neural language models}.
\newblock In \emph{Proc. of ACL}, pages 1828--1843, Online. Association for
  Computational Linguistics.

\bibitem[{Forbes et~al.(2019)Forbes, Holtzman, and
  Choi}]{forbesNeuralLanguageRepresentations2019}
Maxwell Forbes, Ari Holtzman, and Yejin Choi. 2019.
\newblock \href {https://arxiv.org/abs/1908.02899} {Do {{Neural Language
  Representations Learn Physical Commonsense}}?}
\newblock \emph{ArXiv preprint}, abs/1908.02899.

\bibitem[{Gao et~al.(2021)Gao, Fisch, and Chen}]{gao2020making}
Tianyu Gao, Adam Fisch, and Danqi Chen. 2021.
\newblock \href {https://doi.org/10.18653/v1/2021.acl-long.295} {Making
  pre-trained language models better few-shot learners}.
\newblock In \emph{Proc. of ACL}, pages 3816--3830, Online. Association for
  Computational Linguistics.

\bibitem[{Geva et~al.(2021)Geva, Schuster, Berant, and
  Levy}]{geva2020transformer}
Mor Geva, Roei Schuster, Jonathan Berant, and Omer Levy. 2021.
\newblock \href {https://doi.org/10.18653/v1/2021.emnlp-main.446} {Transformer
  feed-forward layers are key-value memories}.
\newblock In \emph{Proc. of EMNLP}, pages 5484--5495, Online and Punta Cana,
  Dominican Republic. Association for Computational Linguistics.

\bibitem[{Gillick et~al.(2019)Gillick, Kulkarni, Lansing, Presta, Baldridge,
  Ie, and Garcia-Olano}]{gillick2019learning}
Daniel Gillick, Sayali Kulkarni, Larry Lansing, Alessandro Presta, Jason
  Baldridge, Eugene Ie, and Diego Garcia-Olano. 2019.
\newblock \href {https://doi.org/10.18653/v1/K19-1049} {Learning dense
  representations for entity retrieval}.
\newblock In \emph{Proceedings of the 23rd Conference on Computational Natural
  Language Learning (CoNLL)}, pages 528--537, Hong Kong, China. Association for
  Computational Linguistics.

\bibitem[{Goldberg(2019)}]{goldbergAssessingBERTSyntactic2019}
Yoav Goldberg. 2019.
\newblock \href {https://arxiv.org/abs/1901.05287} {Assessing {{BERT}}'s
  {{Syntactic Abilities}}}.
\newblock \emph{ArXiv preprint}, abs/1901.05287.

\bibitem[{Guan et~al.(2020)Guan, Huang, Zhao, Zhu, and
  Huang}]{guan2020knowledge}
Jian Guan, Fei Huang, Zhihao Zhao, Xiaoyan Zhu, and Minlie Huang. 2020.
\newblock \href {https://doi.org/10.1162/tacl_a_00302} {A knowledge-enhanced
  pretraining model for commonsense story generation}.
\newblock \emph{Transactions of the Association for Computational Linguistics},
  8:93--108.

\bibitem[{Gupta et~al.(2015)Gupta, Boleda, Baroni, and
  Pad{\'o}}]{gupta-etal-2015-distributional}
Abhijeet Gupta, Gemma Boleda, Marco Baroni, and Sebastian Pad{\'o}. 2015.
\newblock \href {https://doi.org/10.18653/v1/D15-1002} {Distributional vectors
  encode referential attributes}.
\newblock In \emph{Proc. of EMNLP}, pages 12--21, Lisbon, Portugal. Association
  for Computational Linguistics.

\bibitem[{Guu et~al.(2020)Guu, Lee, Tung, Pasupat, and
  Chang}]{DBLP:journals/corr/abs-2002-08909}
Kelvin Guu, Kenton Lee, Zora Tung, Panupong Pasupat, and Ming{-}Wei Chang.
  2020.
\newblock \href {https://arxiv.org/abs/2002.08909} {{REALM:}
  retrieval-augmented language model pre-training}.
\newblock \emph{ArXiv preprint}, abs/2002.08909.

\bibitem[{Ha et~al.(2017)Ha, Dai, and Le}]{ha2016hypernetworks}
David Ha, Andrew~M. Dai, and Quoc~V. Le. 2017.
\newblock \href {https://openreview.net/forum?id=rkpACe1lx} {Hypernetworks}.
\newblock In \emph{Proc. of ICLR}. OpenReview.net.

\bibitem[{Hambardzumyan et~al.(2021)Hambardzumyan, Khachatrian, and
  May}]{hambardzumyan2021warp}
Karen Hambardzumyan, Hrant Khachatrian, and Jonathan May. 2021.
\newblock \href {https://doi.org/10.18653/v1/2021.acl-long.381} {{WARP}:
  {W}ord-level {A}dversarial {R}e{P}rogramming}.
\newblock In \emph{Proc. of ACL}, pages 4921--4933, Online. Association for
  Computational Linguistics.

\bibitem[{Han et~al.(2021)Han, Zhao, Ding, Liu, and Sun}]{han2021ptr}
Xu~Han, Weilin Zhao, Ning Ding, Zhiyuan Liu, and Maosong Sun. 2021.
\newblock \href {https://arxiv.org/abs/2105.11259} {Ptr: Prompt tuning with
  rules for text classification}.
\newblock \emph{ArXiv preprint}, abs/2105.11259.

\bibitem[{Hardt et~al.(2016)Hardt, Price, and
  Srebro}]{DBLP:conf/nips/HardtPNS16}
Moritz Hardt, Eric Price, and Nati Srebro. 2016.
\newblock \href
  {https://proceedings.neurips.cc/paper/2016/hash/9d2682367c3935defcb1f9e247a97c0d-Abstract.html}
  {Equality of opportunity in supervised learning}.
\newblock In \emph{Advances in Neural Information Processing Systems 29: Annual
  Conference on Neural Information Processing Systems 2016, December 5-10,
  2016, Barcelona, Spain}, pages 3315--3323.

\bibitem[{Hase et~al.(2021)Hase, Diab, Celikyilmaz, Li, Kozareva, Stoyanov,
  Bansal, and Iyer}]{hase2021language}
Peter Hase, Mona Diab, Asli Celikyilmaz, Xian Li, Zornitsa Kozareva, Veselin
  Stoyanov, Mohit Bansal, and Srinivasan Iyer. 2021.
\newblock \href {https://arxiv.org/abs/2111.13654} {Do language models have
  beliefs? methods for detecting, updating, and visualizing model beliefs}.
\newblock \emph{ArXiv preprint}, abs/2111.13654.

\bibitem[{Haviv et~al.(2021)Haviv, Berant, and
  Globerson}]{haviv-etal-2021-bertese}
Adi Haviv, Jonathan Berant, and Amir Globerson. 2021.
\newblock \href {https://doi.org/10.18653/v1/2021.eacl-main.316} {{BERT}ese:
  Learning to speak to {BERT}}.
\newblock In \emph{Proceedings of the 16th Conference of the European Chapter
  of the Association for Computational Linguistics: Main Volume}, pages
  3618--3623, Online. Association for Computational Linguistics.

\bibitem[{Heinzerling and Inui(2021)}]{heinzerling2020language}
Benjamin Heinzerling and Kentaro Inui. 2021.
\newblock \href {https://doi.org/10.18653/v1/2021.eacl-main.153} {Language
  models as knowledge bases: On entity representations, storage capacity, and
  paraphrased queries}.
\newblock In \emph{Proceedings of the 16th Conference of the European Chapter
  of the Association for Computational Linguistics: Main Volume}, pages
  1772--1791, Online. Association for Computational Linguistics.

\bibitem[{Hewitt and Manning(2019)}]{hewittStructuralProbeFinding}
John Hewitt and Christopher~D. Manning. 2019.
\newblock \href {https://doi.org/10.18653/v1/N19-1419} {{A} structural probe
  for finding syntax in word representations}.
\newblock In \emph{Proc. of NAACL-HLT}, pages 4129--4138, Minneapolis,
  Minnesota. Association for Computational Linguistics.

\bibitem[{Hochreiter and Schmidhuber(1997)}]{hochreiter1997long}
Sepp Hochreiter and J{\"u}rgen Schmidhuber. 1997.
\newblock Long short-term memory.
\newblock \emph{Neural computation}, 9(8):1735--1780.

\bibitem[{Htut et~al.(2019)Htut, Phang, Bordia, and
  Bowman}]{htutAttentionHeadsBERT2019}
Phu~Mon Htut, Jason Phang, Shikha Bordia, and Samuel~R. Bowman. 2019.
\newblock \href {https://arxiv.org/abs/1911.12246} {Do {{Attention Heads}} in
  {{BERT Track Syntactic Dependencies}}?}
\newblock \emph{ArXiv preprint}, abs/1911.12246.

\bibitem[{Hu et~al.(2022)Hu, Ding, Wang, Liu, Wang, Li, Wu, and
  Sun}]{hu2021knowledgeable}
Shengding Hu, Ning Ding, Huadong Wang, Zhiyuan Liu, Jingang Wang, Juanzi Li,
  Wei Wu, and Maosong Sun. 2022.
\newblock \href {https://doi.org/10.18653/v1/2022.acl-long.158} {Knowledgeable
  prompt-tuning: Incorporating knowledge into prompt verbalizer for text
  classification}.
\newblock In \emph{Proc. of ACL}, pages 2225--2240, Dublin, Ireland.
  Association for Computational Linguistics.

\bibitem[{Jang et~al.(2022)Jang, Ye, and Seo}]{jang2022can}
Joel Jang, Seonghyeon Ye, and Minjoon Seo. 2022.
\newblock \href {https://arxiv.org/abs/2209.12711} {Can large language models
  truly understand prompts? a case study with negated prompts}.
\newblock \emph{ArXiv preprint}, abs/2209.12711.

\bibitem[{Jiang et~al.(2020{\natexlab{a}})Jiang, Anastasopoulos, Araki, Ding,
  and Neubig}]{jiangXFACTRMultilingualFactual2020}
Zhengbao Jiang, Antonios Anastasopoulos, Jun Araki, Haibo Ding, and Graham
  Neubig. 2020{\natexlab{a}}.
\newblock \href {https://doi.org/10.18653/v1/2020.emnlp-main.479} {{X}-{FACTR}:
  Multilingual factual knowledge retrieval from pretrained language models}.
\newblock In \emph{Proc. of EMNLP}, pages 5943--5959, Online. Association for
  Computational Linguistics.

\bibitem[{Jiang et~al.(2020{\natexlab{b}})Jiang, Xu, Araki, and
  Neubig}]{jiangHowCanWe2020}
Zhengbao Jiang, Frank~F. Xu, Jun Araki, and Graham Neubig. 2020{\natexlab{b}}.
\newblock \href {https://doi.org/10.1162/tacl_a_00324} {How can we know what
  language models know?}
\newblock \emph{Transactions of the Association for Computational Linguistics},
  8:423--438.

\bibitem[{Juneja and Agarwal(2022)}]{juneja2022finding}
Jeevesh Juneja and Ritu Agarwal. 2022.
\newblock \href {https://arxiv.org/abs/2205.01366} {Finding patterns in
  knowledge attribution for transformers}.
\newblock \emph{ArXiv preprint}, abs/2205.01366.

\bibitem[{Kassner et~al.(2021)Kassner, Dufter, and
  Sch{\"u}tze}]{kassner2021multilingual}
Nora Kassner, Philipp Dufter, and Hinrich Sch{\"u}tze. 2021.
\newblock \href {https://doi.org/10.18653/v1/2021.eacl-main.284} {Multilingual
  {LAMA}: Investigating knowledge in multilingual pretrained language models}.
\newblock In \emph{Proceedings of the 16th Conference of the European Chapter
  of the Association for Computational Linguistics: Main Volume}, pages
  3250--3258, Online. Association for Computational Linguistics.

\bibitem[{Kassner and Sch{\"u}tze(2020)}]{kassnerNegatedMisprimedProbes2020b}
Nora Kassner and Hinrich Sch{\"u}tze. 2020.
\newblock \href {https://doi.org/10.18653/v1/2020.acl-main.698} {Negated and
  misprimed probes for pretrained language models: Birds can talk, but cannot
  fly}.
\newblock In \emph{Proc. of ACL}, pages 7811--7818, Online. Association for
  Computational Linguistics.

\bibitem[{Ke et~al.(2020)Ke, Ji, Liu, Zhu, and Huang}]{ke2019sentilare}
Pei Ke, Haozhe Ji, Siyang Liu, Xiaoyan Zhu, and Minlie Huang. 2020.
\newblock \href {https://doi.org/10.18653/v1/2020.emnlp-main.567}
  {{S}enti{LARE}: Sentiment-aware language representation learning with
  linguistic knowledge}.
\newblock In \emph{Proc. of EMNLP}, pages 6975--6988, Online. Association for
  Computational Linguistics.

\bibitem[{Kilbertus et~al.(2017)Kilbertus, Rojas{-}Carulla, Parascandolo,
  Hardt, Janzing, and Sch{\"{o}}lkopf}]{DBLP:conf/nips/KilbertusRPHJS17}
Niki Kilbertus, Mateo Rojas{-}Carulla, Giambattista Parascandolo, Moritz Hardt,
  Dominik Janzing, and Bernhard Sch{\"{o}}lkopf. 2017.
\newblock \href
  {https://proceedings.neurips.cc/paper/2017/hash/f5f8590cd58a54e94377e6ae2eded4d9-Abstract.html}
  {Avoiding discrimination through causal reasoning}.
\newblock In \emph{Advances in Neural Information Processing Systems 30: Annual
  Conference on Neural Information Processing Systems 2017, December 4-9, 2017,
  Long Beach, CA, {USA}}, pages 656--666.

\bibitem[{K{\"o}hn(2015)}]{kohn-2015-whats}
Arne K{\"o}hn. 2015.
\newblock \href {https://doi.org/10.18653/v1/D15-1246} {What{'}s in an
  embedding? analyzing word embeddings through multilingual evaluation}.
\newblock In \emph{Proc. of EMNLP}, pages 2067--2073, Lisbon, Portugal.
  Association for Computational Linguistics.

\bibitem[{Kusner et~al.(2017)Kusner, Loftus, Russell, and
  Silva}]{DBLP:conf/nips/KusnerLRS17}
Matt~J. Kusner, Joshua~R. Loftus, Chris Russell, and Ricardo Silva. 2017.
\newblock \href
  {https://proceedings.neurips.cc/paper/2017/hash/a486cd07e4ac3d270571622f4f316ec5-Abstract.html}
  {Counterfactual fairness}.
\newblock In \emph{Advances in Neural Information Processing Systems 30: Annual
  Conference on Neural Information Processing Systems 2017, December 4-9, 2017,
  Long Beach, CA, {USA}}, pages 4066--4076.

\bibitem[{Lampinen et~al.(2022)Lampinen, Dasgupta, Chan, Matthewson, Tessler,
  Creswell, McClelland, Wang, and Hill}]{lampinen2022can}
Andrew~K Lampinen, Ishita Dasgupta, Stephanie~CY Chan, Kory Matthewson,
  Michael~Henry Tessler, Antonia Creswell, James~L McClelland, Jane~X Wang, and
  Felix Hill. 2022.
\newblock \href {https://arxiv.org/abs/2204.02329} {Can language models learn
  from explanations in context?}
\newblock \emph{ArXiv preprint}, abs/2204.02329.

\bibitem[{Lan et~al.(2020)Lan, Chen, Goodman, Gimpel, Sharma, and
  Soricut}]{lan2019albert}
Zhenzhong Lan, Mingda Chen, Sebastian Goodman, Kevin Gimpel, Piyush Sharma, and
  Radu Soricut. 2020.
\newblock \href {https://openreview.net/forum?id=H1eA7AEtvS} {{ALBERT:} {A}
  lite {BERT} for self-supervised learning of language representations}.
\newblock In \emph{Proc. of ICLR}. OpenReview.net.

\bibitem[{Lauscher et~al.(2020)Lauscher, Vuli{\'c}, Ponti, Korhonen, and
  Glava{\v{s}}}]{lauscher2019specializing}
Anne Lauscher, Ivan Vuli{\'c}, Edoardo~Maria Ponti, Anna Korhonen, and Goran
  Glava{\v{s}}. 2020.
\newblock \href {https://doi.org/10.18653/v1/2020.coling-main.118}
  {Specializing unsupervised pretraining models for word-level semantic
  similarity}.
\newblock In \emph{Proceedings of the 28th International Conference on
  Computational Linguistics}, pages 1371--1383, Barcelona, Spain (Online).
  International Committee on Computational Linguistics.

\bibitem[{Lee et~al.(2022)Lee, Kadakia, Tan, Agarwal, Feng, Shibuya, Mitani,
  Sekiya, Pujara, and Ren}]{lee2021good}
Dong-Ho Lee, Akshen Kadakia, Kangmin Tan, Mahak Agarwal, Xinyu Feng, Takashi
  Shibuya, Ryosuke Mitani, Toshiyuki Sekiya, Jay Pujara, and Xiang Ren. 2022.
\newblock \href {https://doi.org/10.18653/v1/2022.acl-long.192} {Good examples
  make a faster learner: Simple demonstration-based learning for low-resource
  {NER}}.
\newblock In \emph{Proc. of ACL}, pages 2687--2700, Dublin, Ireland.
  Association for Computational Linguistics.

\bibitem[{Lester et~al.(2021)Lester, Al-Rfou, and Constant}]{lester2021power}
Brian Lester, Rami Al-Rfou, and Noah Constant. 2021.
\newblock \href {https://doi.org/10.18653/v1/2021.emnlp-main.243} {The power of
  scale for parameter-efficient prompt tuning}.
\newblock In \emph{Proc. of EMNLP}, pages 3045--3059, Online and Punta Cana,
  Dominican Republic. Association for Computational Linguistics.

\bibitem[{Levine et~al.(2020)Levine, Lenz, Dagan, Ram, Padnos, Sharir,
  Shalev-Shwartz, Shashua, and Shoham}]{levine2019sensebert}
Yoav Levine, Barak Lenz, Or~Dagan, Ori Ram, Dan Padnos, Or~Sharir, Shai
  Shalev-Shwartz, Amnon Shashua, and Yoav Shoham. 2020.
\newblock \href {https://doi.org/10.18653/v1/2020.acl-main.423} {{S}ense{BERT}:
  Driving some sense into {BERT}}.
\newblock In \emph{Proc. of ACL}, pages 4656--4667, Online. Association for
  Computational Linguistics.

\bibitem[{Lewis et~al.(2020{\natexlab{a}})Lewis, Liu, Goyal, Ghazvininejad,
  Mohamed, Levy, Stoyanov, and Zettlemoyer}]{lewis2019bart}
Mike Lewis, Yinhan Liu, Naman Goyal, Marjan Ghazvininejad, Abdelrahman Mohamed,
  Omer Levy, Veselin Stoyanov, and Luke Zettlemoyer. 2020{\natexlab{a}}.
\newblock \href {https://doi.org/10.18653/v1/2020.acl-main.703} {{BART}:
  Denoising sequence-to-sequence pre-training for natural language generation,
  translation, and comprehension}.
\newblock In \emph{Proc. of ACL}, pages 7871--7880, Online. Association for
  Computational Linguistics.

\bibitem[{Lewis et~al.(2020{\natexlab{b}})Lewis, Perez, Piktus, Petroni,
  Karpukhin, Goyal, K{\"{u}}ttler, Lewis, Yih, Rockt{\"{a}}schel, Riedel, and
  Kiela}]{DBLP:conf/nips/LewisPPPKGKLYR020}
Patrick S.~H. Lewis, Ethan Perez, Aleksandra Piktus, Fabio Petroni, Vladimir
  Karpukhin, Naman Goyal, Heinrich K{\"{u}}ttler, Mike Lewis, Wen{-}tau Yih,
  Tim Rockt{\"{a}}schel, Sebastian Riedel, and Douwe Kiela. 2020{\natexlab{b}}.
\newblock \href
  {https://proceedings.neurips.cc/paper/2020/hash/6b493230205f780e1bc26945df7481e5-Abstract.html}
  {Retrieval-augmented generation for knowledge-intensive {NLP} tasks}.
\newblock In \emph{Advances in Neural Information Processing Systems 33: Annual
  Conference on Neural Information Processing Systems 2020, NeurIPS 2020,
  December 6-12, 2020, virtual}.

\bibitem[{Li et~al.(2022{\natexlab{a}})Li, Li, Shang, Dong, Sun, Liu, Ji,
  Jiang, and Liu}]{li-etal-2022-pre}
Shaobo Li, Xiaoguang Li, Lifeng Shang, Zhenhua Dong, Chengjie Sun, Bingquan
  Liu, Zhenzhou Ji, Xin Jiang, and Qun Liu. 2022{\natexlab{a}}.
\newblock \href {https://doi.org/10.18653/v1/2022.findings-acl.136} {How
  pre-trained language models capture factual knowledge? a causal-inspired
  analysis}.
\newblock In \emph{Findings of the Association for Computational Linguistics:
  ACL 2022}, pages 1720--1732, Dublin, Ireland. Association for Computational
  Linguistics.

\bibitem[{Li et~al.(2022{\natexlab{b}})Li, Chen, Shen, Chen, Zhang, Li, Wang,
  Qian, Peng, Mao et~al.}]{li2022explanations}
Shiyang Li, Jianshu Chen, Yelong Shen, Zhiyu Chen, Xinlu Zhang, Zekun Li, Hong
  Wang, Jing Qian, Baolin Peng, Yi~Mao, et~al. 2022{\natexlab{b}}.
\newblock \href {https://arxiv.org/abs/2210.06726} {Explanations from large
  language models make small reasoners better}.
\newblock \emph{ArXiv preprint}, abs/2210.06726.

\bibitem[{Li and
  Liang(2021{\natexlab{a}})}]{liPrefixTuningOptimizingContinuous2021}
Xiang~Lisa Li and Percy Liang. 2021{\natexlab{a}}.
\newblock \href {https://doi.org/10.18653/v1/2021.acl-long.353} {Prefix-tuning:
  Optimizing continuous prompts for generation}.
\newblock In \emph{Proc. of ACL}, pages 4582--4597, Online. Association for
  Computational Linguistics.

\bibitem[{Li and Liang(2021{\natexlab{b}})}]{li2021prefix}
Xiang~Lisa Li and Percy Liang. 2021{\natexlab{b}}.
\newblock \href {https://doi.org/10.18653/v1/2021.acl-long.353} {Prefix-tuning:
  Optimizing continuous prompts for generation}.
\newblock In \emph{Proc. of ACL}, pages 4582--4597, Online. Association for
  Computational Linguistics.

\bibitem[{Lin et~al.(2020{\natexlab{a}})Lin, Lee, Khanna, and
  Ren}]{linBirdsHaveFour2020}
Bill~Yuchen Lin, Seyeon Lee, Rahul Khanna, and Xiang Ren. 2020{\natexlab{a}}.
\newblock \href {https://doi.org/10.18653/v1/2020.emnlp-main.557} {{B}irds have
  four legs?! {N}umer{S}ense: {P}robing {N}umerical {C}ommonsense {K}nowledge
  of {P}re-{T}rained {L}anguage {M}odels}.
\newblock In \emph{Proc. of EMNLP}, pages 6862--6868, Online. Association for
  Computational Linguistics.

\bibitem[{Lin et~al.(2020{\natexlab{b}})Lin, Lee, Khanna, and
  Ren}]{lin2020birds}
Bill~Yuchen Lin, Seyeon Lee, Rahul Khanna, and Xiang Ren. 2020{\natexlab{b}}.
\newblock \href {https://doi.org/10.18653/v1/2020.emnlp-main.557} {{B}irds have
  four legs?! {N}umer{S}ense: {P}robing {N}umerical {C}ommonsense {K}nowledge
  of {P}re-{T}rained {L}anguage {M}odels}.
\newblock In \emph{Proc. of EMNLP}, pages 6862--6868, Online. Association for
  Computational Linguistics.

\bibitem[{Lin et~al.(2019)Lin, Tan, and Frank}]{linOpenSesameGetting2019}
Yongjie Lin, Yi~Chern Tan, and Robert Frank. 2019.
\newblock \href {https://doi.org/10.18653/v1/W19-4825} {Open sesame: Getting
  inside {BERT}{'}s linguistic knowledge}.
\newblock In \emph{Proceedings of the 2019 ACL Workshop BlackboxNLP: Analyzing
  and Interpreting Neural Networks for NLP}, pages 241--253, Florence, Italy.
  Association for Computational Linguistics.

\bibitem[{Liu et~al.(2022)Liu, Shen, Zhang, Dolan, Carin, and
  Chen}]{liu2021makes}
Jiachang Liu, Dinghan Shen, Yizhe Zhang, Bill Dolan, Lawrence Carin, and Weizhu
  Chen. 2022.
\newblock \href {https://doi.org/10.18653/v1/2022.deelio-1.10} {What makes good
  in-context examples for {GPT}-3?}
\newblock In \emph{Proceedings of Deep Learning Inside Out (DeeLIO 2022): The
  3rd Workshop on Knowledge Extraction and Integration for Deep Learning
  Architectures}, pages 100--114, Dublin, Ireland and Online. Association for
  Computational Linguistics.

\bibitem[{Liu et~al.(2019{\natexlab{a}})Liu, Gardner, Belinkov, Peters, and
  Smith}]{liu-etal-2019-linguistic}
Nelson~F. Liu, Matt Gardner, Yonatan Belinkov, Matthew~E. Peters, and Noah~A.
  Smith. 2019{\natexlab{a}}.
\newblock \href {https://doi.org/10.18653/v1/N19-1112} {Linguistic knowledge
  and transferability of contextual representations}.
\newblock In \emph{Proc. of NAACL-HLT}, pages 1073--1094, Minneapolis,
  Minnesota. Association for Computational Linguistics.

\bibitem[{Liu et~al.(2019{\natexlab{b}})Liu, Gardner, Belinkov, Peters, and
  Smith}]{liuLinguisticKnowledgeTransferability2019}
Nelson~F. Liu, Matt Gardner, Yonatan Belinkov, Matthew~E. Peters, and Noah~A.
  Smith. 2019{\natexlab{b}}.
\newblock \href {https://doi.org/10.18653/v1/N19-1112} {Linguistic knowledge
  and transferability of contextual representations}.
\newblock In \emph{Proc. of NAACL-HLT}, pages 1073--1094, Minneapolis,
  Minnesota. Association for Computational Linguistics.

\bibitem[{Liu et~al.(2021{\natexlab{a}})Liu, Yuan, Fu, Jiang, Hayashi, and
  Neubig}]{liu2021pre}
Pengfei Liu, Weizhe Yuan, Jinlan Fu, Zhengbao Jiang, Hiroaki Hayashi, and
  Graham Neubig. 2021{\natexlab{a}}.
\newblock \href {https://arxiv.org/abs/2107.13586} {Pre-train, prompt, and
  predict: A systematic survey of prompting methods in natural language
  processing}.
\newblock \emph{ArXiv preprint}, abs/2107.13586.

\bibitem[{Liu et~al.(2020)Liu, Zhou, Zhao, Wang, Ju, Deng, and Wang}]{liu2020k}
Weijie Liu, Peng Zhou, Zhe Zhao, Zhiruo Wang, Qi~Ju, Haotang Deng, and Ping
  Wang. 2020.
\newblock \href {https://aaai.org/ojs/index.php/AAAI/article/view/5681}
  {{K-BERT:} enabling language representation with knowledge graph}.
\newblock In \emph{The Thirty-Fourth {AAAI} Conference on Artificial
  Intelligence, {AAAI} 2020, The Thirty-Second Innovative Applications of
  Artificial Intelligence Conference, {IAAI} 2020, The Tenth {AAAI} Symposium
  on Educational Advances in Artificial Intelligence, {EAAI} 2020, New York,
  NY, USA, February 7-12, 2020}, pages 2901--2908. {AAAI} Press.

\bibitem[{Liu et~al.(2021{\natexlab{b}})Liu, Zheng, Du, Ding, Qian, Yang, and
  Tang}]{liu2021gpt}
Xiao Liu, Yanan Zheng, Zhengxiao Du, Ming Ding, Yujie Qian, Zhilin Yang, and
  Jie Tang. 2021{\natexlab{b}}.
\newblock \href {https://arxiv.org/abs/2103.10385} {Gpt understands, too}.
\newblock \emph{ArXiv preprint}, abs/2103.10385.

\bibitem[{Liu et~al.(2019{\natexlab{c}})Liu, Ott, Goyal, Du, Joshi, Chen, Levy,
  Lewis, Zettlemoyer, and Stoyanov}]{liu2019roberta}
Yinhan Liu, Myle Ott, Naman Goyal, Jingfei Du, Mandar Joshi, Danqi Chen, Omer
  Levy, Mike Lewis, Luke Zettlemoyer, and Veselin Stoyanov. 2019{\natexlab{c}}.
\newblock \href {https://arxiv.org/abs/1907.11692} {Roberta: A robustly
  optimized bert pretraining approach}.
\newblock \emph{ArXiv preprint}, abs/1907.11692.

\bibitem[{Liu et~al.(2021{\natexlab{c}})Liu, Wang, Kasai, Hajishirzi, and
  Smith}]{liu-etal-2021-probing-across}
Zeyu Liu, Yizhong Wang, Jungo Kasai, Hannaneh Hajishirzi, and Noah~A. Smith.
  2021{\natexlab{c}}.
\newblock \href {https://doi.org/10.18653/v1/2021.findings-emnlp.71} {Probing
  across time: What does {R}o{BERT}a know and when?}
\newblock In \emph{Findings of the Association for Computational Linguistics:
  EMNLP 2021}, pages 820--842, Punta Cana, Dominican Republic. Association for
  Computational Linguistics.

\bibitem[{Logeswaran et~al.(2019)Logeswaran, Chang, Lee, Toutanova, Devlin, and
  Lee}]{logeswaran2019zero}
Lajanugen Logeswaran, Ming-Wei Chang, Kenton Lee, Kristina Toutanova, Jacob
  Devlin, and Honglak Lee. 2019.
\newblock \href {https://doi.org/10.18653/v1/P19-1335} {Zero-shot entity
  linking by reading entity descriptions}.
\newblock In \emph{Proc. of ACL}, pages 3449--3460, Florence, Italy.
  Association for Computational Linguistics.

\bibitem[{Lu et~al.(2022)Lu, Bartolo, Moore, Riedel, and
  Stenetorp}]{lu2021fantastically}
Yao Lu, Max Bartolo, Alastair Moore, Sebastian Riedel, and Pontus Stenetorp.
  2022.
\newblock \href {https://doi.org/10.18653/v1/2022.acl-long.556} {Fantastically
  ordered prompts and where to find them: Overcoming few-shot prompt order
  sensitivity}.
\newblock In \emph{Proc. of ACL}, pages 8086--8098, Dublin, Ireland.
  Association for Computational Linguistics.

\bibitem[{Ma et~al.(2021)Ma, Ilievski, Francis, Bisk, Nyberg, and
  Oltramari}]{ma2021knowledge}
Kaixin Ma, Filip Ilievski, Jonathan Francis, Yonatan Bisk, Eric Nyberg, and
  Alessandro Oltramari. 2021.
\newblock Knowledge-driven data construction for zero-shot evaluation in
  commonsense question answering.
\newblock In \emph{Proc. of AAAI}, volume~35, pages 13507--13515.

\bibitem[{Madaan et~al.(2022)Madaan, Tandon, Clark, and
  Yang}]{madaan2022memory}
Aman Madaan, Niket Tandon, Peter Clark, and Yiming Yang. 2022.
\newblock \href {https://arxiv.org/abs/2201.06009} {Memory-assisted prompt
  editing to improve gpt-3 after deployment}.
\newblock \emph{ArXiv preprint}, abs/2201.06009.

\bibitem[{Manning et~al.(2020)Manning, Clark, Hewitt, Khandelwal, and
  Levy}]{manning2020emergent}
Christopher~D Manning, Kevin Clark, John Hewitt, Urvashi Khandelwal, and Omer
  Levy. 2020.
\newblock Emergent linguistic structure in artificial neural networks trained
  by self-supervision.
\newblock \emph{Proceedings of the National Academy of Sciences},
  117(48):30046--30054.

\bibitem[{Meng et~al.(2022)Meng, Bau, Andonian, and
  Belinkov}]{meng2022locating}
Kevin Meng, David Bau, Alex Andonian, and Yonatan Belinkov. 2022.
\newblock \href {https://arxiv.org/abs/2202.05262} {Locating and editing
  factual knowledge in gpt}.
\newblock \emph{ArXiv preprint}, abs/2202.05262.

\bibitem[{Miller(1992)}]{miller1995wordnet}
George~A. Miller. 1992.
\newblock \href {https://aclanthology.org/H92-1116} {{W}ord{N}et: A lexical
  database for {E}nglish}.
\newblock In \emph{Speech and Natural Language: Proceedings of a Workshop Held
  at Harriman, New York, {F}ebruary 23-26, 1992}.

\bibitem[{Min et~al.(2022)Min, Lyu, Holtzman, Artetxe, Lewis, Hajishirzi, and
  Zettlemoyer}]{min2022rethinking}
Sewon Min, Xinxi Lyu, Ari Holtzman, Mikel Artetxe, Mike Lewis, Hannaneh
  Hajishirzi, and Luke Zettlemoyer. 2022.
\newblock \href {https://arxiv.org/abs/2202.12837} {Rethinking the role of
  demonstrations: What makes in-context learning work?}
\newblock \emph{ArXiv preprint}, abs/2202.12837.

\bibitem[{Mitchell et~al.(2021)Mitchell, Lin, Bosselut, Finn, and
  Manning}]{mitchell2021fast}
Eric Mitchell, Charles Lin, Antoine Bosselut, Chelsea Finn, and Christopher~D
  Manning. 2021.
\newblock \href {https://arxiv.org/abs/2110.11309} {Fast model editing at
  scale}.
\newblock \emph{ArXiv preprint}, abs/2110.11309.

\bibitem[{Mitchell et~al.(2022)Mitchell, Lin, Bosselut, Manning, and
  Finn}]{mitchell2022memory}
Eric Mitchell, Charles Lin, Antoine Bosselut, Christopher~D Manning, and
  Chelsea Finn. 2022.
\newblock Memory-based model editing at scale.
\newblock In \emph{International Conference on Machine Learning}, pages
  15817--15831. PMLR.

\bibitem[{Navigli and Ponzetto(2010)}]{navigli2010babelnet}
Roberto Navigli and Simone~Paolo Ponzetto. 2010.
\newblock \href {https://aclanthology.org/P10-1023} {{B}abel{N}et: Building a
  very large multilingual semantic network}.
\newblock In \emph{Proc. of ACL}, pages 216--225, Uppsala, Sweden. Association
  for Computational Linguistics.

\bibitem[{Nilson(1974)}]{Nilson:ai:1974}
Nils~J. Nilson. 1974.
\newblock Artificial intelligence.
\newblock In \emph{Information Processing}, pages 778--801.

\bibitem[{Ouyang et~al.(2022)Ouyang, Wu, Jiang, Almeida, Wainwright, Mishkin,
  Zhang, Agarwal, Slama, Ray et~al.}]{ouyang2022training}
Long Ouyang, Jeff Wu, Xu~Jiang, Diogo Almeida, Carroll~L Wainwright, Pamela
  Mishkin, Chong Zhang, Sandhini Agarwal, Katarina Slama, Alex Ray, et~al.
  2022.
\newblock \href {https://arxiv.org/abs/2203.02155} {Training language models to
  follow instructions with human feedback}.
\newblock \emph{ArXiv preprint}, abs/2203.02155.

\bibitem[{Ozturkler et~al.(2022)Ozturkler, Malkin, Wang, and
  Jojic}]{ozturkler2022thinksum}
Batu Ozturkler, Nikolay Malkin, Zhen Wang, and Nebojsa Jojic. 2022.
\newblock \href {https://arxiv.org/abs/2210.01293} {Thinksum: Probabilistic
  reasoning over sets using large language models}.
\newblock \emph{ArXiv preprint}, abs/2210.01293.

\bibitem[{P{\'e}rez et~al.(2009)P{\'e}rez, Arenas, and
  Gutierrez}]{perez2009semantics}
Jorge P{\'e}rez, Marcelo Arenas, and Claudio Gutierrez. 2009.
\newblock Semantics and complexity of sparql.
\newblock \emph{ACM Transactions on Database Systems (TODS)}, 34(3):1--45.

\bibitem[{P{\'e}rez-Mayos et~al.(2021)P{\'e}rez-Mayos, Ballesteros, and
  Wanner}]{perez2021much}
Laura P{\'e}rez-Mayos, Miguel Ballesteros, and Leo Wanner. 2021.
\newblock \href {https://doi.org/10.18653/v1/2021.emnlp-main.118} {How much
  pretraining data do language models need to learn syntax?}
\newblock In \emph{Proc. of EMNLP}, pages 1571--1582, Online and Punta Cana,
  Dominican Republic. Association for Computational Linguistics.

\bibitem[{Peters et~al.(2019)Peters, Neumann, Logan, Schwartz, Joshi, Singh,
  and Smith}]{peters2019knowledge}
Matthew~E. Peters, Mark Neumann, Robert Logan, Roy Schwartz, Vidur Joshi,
  Sameer Singh, and Noah~A. Smith. 2019.
\newblock \href {https://doi.org/10.18653/v1/D19-1005} {Knowledge enhanced
  contextual word representations}.
\newblock In \emph{Proc. of EMNLP}, pages 43--54, Hong Kong, China. Association
  for Computational Linguistics.

\bibitem[{Petroni et~al.(2019)Petroni, Rockt{\"a}schel, Riedel, Lewis, Bakhtin,
  Wu, and Miller}]{petroniLanguageModelsKnowledge2019}
Fabio Petroni, Tim Rockt{\"a}schel, Sebastian Riedel, Patrick Lewis, Anton
  Bakhtin, Yuxiang Wu, and Alexander Miller. 2019.
\newblock \href {https://doi.org/10.18653/v1/D19-1250} {Language models as
  knowledge bases?}
\newblock In \emph{Proc. of EMNLP}, pages 2463--2473, Hong Kong, China.
  Association for Computational Linguistics.

\bibitem[{Poerner et~al.(2020)Poerner, Waltinger, and
  Sch{\"u}tze}]{poernerEBERTEfficientYetEffectiveEntity2020}
Nina Poerner, Ulli Waltinger, and Hinrich Sch{\"u}tze. 2020.
\newblock \href {https://doi.org/10.18653/v1/2020.findings-emnlp.71}
  {{E}-{BERT}: Efficient-yet-effective entity embeddings for {BERT}}.
\newblock In \emph{Findings of the Association for Computational Linguistics:
  EMNLP 2020}, pages 803--818, Online. Association for Computational
  Linguistics.

\bibitem[{Qin and Eisner(2021)}]{qin2021learning}
Guanghui Qin and Jason Eisner. 2021.
\newblock \href {https://doi.org/10.18653/v1/2021.naacl-main.410} {Learning how
  to ask: Querying {LM}s with mixtures of soft prompts}.
\newblock In \emph{Proceedings of the 2021 Conference of the North American
  Chapter of the Association for Computational Linguistics: Human Language
  Technologies}, pages 5203--5212, Online. Association for Computational
  Linguistics.

\bibitem[{Qin et~al.(2021)Qin, Lin, Takanobu, Liu, Li, Ji, Huang, Sun, and
  Zhou}]{qin2020erica}
Yujia Qin, Yankai Lin, Ryuichi Takanobu, Zhiyuan Liu, Peng Li, Heng Ji, Minlie
  Huang, Maosong Sun, and Jie Zhou. 2021.
\newblock \href {https://doi.org/10.18653/v1/2021.acl-long.260} {{ERICA}:
  Improving entity and relation understanding for pre-trained language models
  via contrastive learning}.
\newblock In \emph{Proc. of ACL}, pages 3350--3363, Online. Association for
  Computational Linguistics.

\bibitem[{Qiu et~al.(2020)Qiu, Sun, Xu, Shao, Dai, and
  Huang}]{xipengqiuPretrainedModelsNatural2020}
Xipeng Qiu, Tianxiang Sun, Yige Xu, Yunfan Shao, Ning Dai, and Xuanjing Huang.
  2020.
\newblock \href {https://arxiv.org/abs/2003.08271} {Pre-trained models for
  natural language processing: {A} survey}.
\newblock \emph{ArXiv preprint}, abs/2003.08271.

\bibitem[{Radford et~al.(2019{\natexlab{a}})Radford, Wu, Child, Luan, Amodei,
  and Sutskever}]{radfordLanguageModelsAre2019}
Alec Radford, Jeffrey Wu, Rewon Child, David Luan, Dario Amodei, and Ilya
  Sutskever. 2019{\natexlab{a}}.
\newblock Language models are unsupervised multitask learners.
\newblock \emph{OpenAI blog}, 1(8):9.

\bibitem[{Radford et~al.(2019{\natexlab{b}})Radford, Wu, Child, Luan, Amodei,
  Sutskever et~al.}]{radford2019language}
Alec Radford, Jeffrey Wu, Rewon Child, David Luan, Dario Amodei, Ilya
  Sutskever, et~al. 2019{\natexlab{b}}.
\newblock Language models are unsupervised multitask learners.
\newblock \emph{OpenAI blog}, 1(8):9.

\bibitem[{Raffel et~al.(2020)Raffel, Shazeer, Roberts, Lee, Narang, Matena,
  Zhou, Li, Liu et~al.}]{raffel2020exploring}
Colin Raffel, Noam Shazeer, Adam Roberts, Katherine Lee, Sharan Narang, Michael
  Matena, Yanqi Zhou, Wei Li, Peter~J Liu, et~al. 2020.
\newblock Exploring the limits of transfer learning with a unified text-to-text
  transformer.
\newblock \emph{J. Mach. Learn. Res.}, 21(140):1--67.

\bibitem[{Raghu et~al.(2017)Raghu, Gilmer, Yosinski, and
  Sohl{-}Dickstein}]{raghu2017svcca}
Maithra Raghu, Justin Gilmer, Jason Yosinski, and Jascha Sohl{-}Dickstein.
  2017.
\newblock \href
  {https://proceedings.neurips.cc/paper/2017/hash/dc6a7e655d7e5840e66733e9ee67cc69-Abstract.html}
  {{SVCCA:} singular vector canonical correlation analysis for deep learning
  dynamics and interpretability}.
\newblock In \emph{Advances in Neural Information Processing Systems 30: Annual
  Conference on Neural Information Processing Systems 2017, December 4-9, 2017,
  Long Beach, CA, {USA}}, pages 6076--6085.

\bibitem[{Razniewski et~al.(2021)Razniewski, Yates, Kassner, and
  Weikum}]{razniewski2021language}
Simon Razniewski, Andrew Yates, Nora Kassner, and Gerhard Weikum. 2021.
\newblock \href {https://arxiv.org/abs/2110.04888} {Language models as or for
  knowledge bases}.
\newblock \emph{ArXiv preprint}, abs/2110.04888.

\bibitem[{Reynolds and McDonell(2021)}]{reynolds2021prompt}
Laria Reynolds and Kyle McDonell. 2021.
\newblock Prompt programming for large language models: Beyond the few-shot
  paradigm.
\newblock In \emph{Extended Abstracts of the 2021 CHI Conference on Human
  Factors in Computing Systems}, pages 1--7.

\bibitem[{Roberts et~al.(2020)Roberts, Raffel, and
  Shazeer}]{robertsHowMuchKnowledge2020a}
Adam Roberts, Colin Raffel, and Noam Shazeer. 2020.
\newblock \href {https://doi.org/10.18653/v1/2020.emnlp-main.437} {How much
  knowledge can you pack into the parameters of a language model?}
\newblock In \emph{Proc. of EMNLP}, pages 5418--5426, Online. Association for
  Computational Linguistics.

\bibitem[{Rogers et~al.(2020)Rogers, Kovaleva, and
  Rumshisky}]{rogers2020primer}
Anna Rogers, Olga Kovaleva, and Anna Rumshisky. 2020.
\newblock \href {https://doi.org/10.1162/tacl_a_00349} {A primer in
  {BERT}ology: What we know about how {BERT} works}.
\newblock \emph{Transactions of the Association for Computational Linguistics},
  8:842--866.

\bibitem[{Rubin et~al.(2022)Rubin, Herzig, and Berant}]{rubin2021learning}
Ohad Rubin, Jonathan Herzig, and Jonathan Berant. 2022.
\newblock \href {https://doi.org/10.18653/v1/2022.naacl-main.191} {Learning to
  retrieve prompts for in-context learning}.
\newblock In \emph{Proceedings of the 2022 Conference of the North American
  Chapter of the Association for Computational Linguistics: Human Language
  Technologies}, pages 2655--2671, Seattle, United States. Association for
  Computational Linguistics.

\bibitem[{Sachan et~al.(2021)Sachan, Zhang, Qi, and
  Hamilton}]{sachan2020syntax}
Devendra Sachan, Yuhao Zhang, Peng Qi, and William~L. Hamilton. 2021.
\newblock \href {https://doi.org/10.18653/v1/2021.eacl-main.228} {Do syntax
  trees help pre-trained transformers extract information?}
\newblock In \emph{Proceedings of the 16th Conference of the European Chapter
  of the Association for Computational Linguistics: Main Volume}, pages
  2647--2661, Online. Association for Computational Linguistics.

\bibitem[{Sanh et~al.(2021)Sanh, Webson, Raffel, Bach, Sutawika, Alyafeai,
  Chaffin, Stiegler, Scao, Raja et~al.}]{sanh2021multitask}
Victor Sanh, Albert Webson, Colin Raffel, Stephen~H Bach, Lintang Sutawika,
  Zaid Alyafeai, Antoine Chaffin, Arnaud Stiegler, Teven~Le Scao, Arun Raja,
  et~al. 2021.
\newblock \href {https://arxiv.org/abs/2110.08207} {Multitask prompted training
  enables zero-shot task generalization}.
\newblock \emph{ArXiv preprint}, abs/2110.08207.

\bibitem[{Saphra and Lopez(2019)}]{saphra2018understanding}
Naomi Saphra and Adam Lopez. 2019.
\newblock \href {https://doi.org/10.18653/v1/N19-1329} {Understanding learning
  dynamics of language models with {SVCCA}}.
\newblock In \emph{Proc. of NAACL-HLT}, pages 3257--3267, Minneapolis,
  Minnesota. Association for Computational Linguistics.

\bibitem[{Saphra and Lopez(2020)}]{saphra-lopez-2020-lstms}
Naomi Saphra and Adam Lopez. 2020.
\newblock \href {https://doi.org/10.18653/v1/2020.findings-emnlp.252} {{LSTM}s
  compose{---}and {L}earn{---}{B}ottom-up}.
\newblock In \emph{Findings of the Association for Computational Linguistics:
  EMNLP 2020}, pages 2797--2809, Online. Association for Computational
  Linguistics.

\bibitem[{Scao et~al.(2022)Scao, Fan, Akiki, Pavlick, Ili{\'c}, Hesslow,
  Castagn{\'e}, Luccioni, Yvon, Gall{\'e} et~al.}]{scao2022bloom}
Teven~Le Scao, Angela Fan, Christopher Akiki, Ellie Pavlick, Suzana Ili{\'c},
  Daniel Hesslow, Roman Castagn{\'e}, Alexandra~Sasha Luccioni, Fran{\c{c}}ois
  Yvon, Matthias Gall{\'e}, et~al. 2022.
\newblock \href {https://arxiv.org/abs/2211.05100} {Bloom: A 176b-parameter
  open-access multilingual language model}.
\newblock \emph{ArXiv preprint}, abs/2211.05100.

\bibitem[{Schick and Sch{\"u}tze(2021)}]{schick2020exploiting}
Timo Schick and Hinrich Sch{\"u}tze. 2021.
\newblock \href {https://doi.org/10.18653/v1/2021.eacl-main.20} {Exploiting
  cloze-questions for few-shot text classification and natural language
  inference}.
\newblock In \emph{Proceedings of the 16th Conference of the European Chapter
  of the Association for Computational Linguistics: Main Volume}, pages
  255--269, Online. Association for Computational Linguistics.

\bibitem[{Schick et~al.(2021)Schick, Udupa, and Sch{\"u}tze}]{schick2021self}
Timo Schick, Sahana Udupa, and Hinrich Sch{\"u}tze. 2021.
\newblock \href {https://doi.org/10.1162/tacl_a_00434} {Self-diagnosis and
  self-debiasing: A proposal for reducing corpus-based bias in {NLP}}.
\newblock \emph{Transactions of the Association for Computational Linguistics},
  9:1408--1424.

\bibitem[{Schreiber et~al.(2000)Schreiber, Schreiber, Akkermans, Anjewierden,
  Shadbolt, de~Hoog, Van~de Velde, and Wielinga}]{schreiber2000knowledge}
August~Th Schreiber, Guus Schreiber, Hans Akkermans, Anjo Anjewierden, Nigel
  Shadbolt, Robert de~Hoog, Walter Van~de Velde, and Bob Wielinga. 2000.
\newblock \emph{Knowledge engineering and management: the CommonKADS
  methodology}.
\newblock MIT press.

\bibitem[{Shen et~al.(2020)Shen, Mao, He, Long, Trischler, and
  Chen}]{shen2020exploiting}
Tao Shen, Yi~Mao, Pengcheng He, Guodong Long, Adam Trischler, and Weizhu Chen.
  2020.
\newblock \href {https://doi.org/10.18653/v1/2020.emnlp-main.722} {Exploiting
  structured knowledge in text via graph-guided representation learning}.
\newblock In \emph{Proc. of EMNLP}, pages 8980--8994, Online. Association for
  Computational Linguistics.

\bibitem[{Shin et~al.(2020)Shin, Razeghi, Logan~IV, Wallace, and
  Singh}]{shinAutoPromptElicitingKnowledge2020}
Taylor Shin, Yasaman Razeghi, Robert~L. Logan~IV, Eric Wallace, and Sameer
  Singh. 2020.
\newblock \href {https://doi.org/10.18653/v1/2020.emnlp-main.346}
  {{A}uto{P}rompt: {E}liciting {K}nowledge from {L}anguage {M}odels with
  {A}utomatically {G}enerated {P}rompts}.
\newblock In \emph{Proc. of EMNLP}, pages 4222--4235, Online. Association for
  Computational Linguistics.

\bibitem[{Shwartz et~al.(2020)Shwartz, West, Le~Bras, Bhagavatula, and
  Choi}]{shwartz2020unsupervised}
Vered Shwartz, Peter West, Ronan Le~Bras, Chandra Bhagavatula, and Yejin Choi.
  2020.
\newblock \href {https://doi.org/10.18653/v1/2020.emnlp-main.373} {Unsupervised
  commonsense question answering with self-talk}.
\newblock In \emph{Proc. of EMNLP}, pages 4615--4629, Online. Association for
  Computational Linguistics.

\bibitem[{Sinitsin et~al.(2020)Sinitsin, Plokhotnyuk, Pyrkin, Popov, and
  Babenko}]{sinitsin2020editable}
Anton Sinitsin, Vsevolod Plokhotnyuk, Dmitriy Pyrkin, Sergei Popov, and Artem
  Babenko. 2020.
\newblock \href {https://openreview.net/forum?id=HJedXaEtvS} {Editable neural
  networks}.
\newblock In \emph{Proc. of ICLR}. OpenReview.net.

\bibitem[{Song et~al.(2022)Song, Liang, Li, Li, Wang, Peng, Wu, and
  Yu}]{song2022improving}
Jian Song, Di~Liang, Rumei Li, Yuntao Li, Sirui Wang, Minlong Peng, Wei Wu, and
  Yongxin Yu. 2022.
\newblock \href {https://arxiv.org/abs/2210.08471} {Improving semantic matching
  through dependency-enhanced pre-trained model with adaptive fusion}.
\newblock \emph{ArXiv preprint}, abs/2210.08471.

\bibitem[{Song et~al.(2019)Song, Tan, Qin, Lu, and Liu}]{song2019mass}
Kaitao Song, Xu~Tan, Tao Qin, Jianfeng Lu, and Tie{-}Yan Liu. 2019.
\newblock \href {http://proceedings.mlr.press/v97/song19d.html} {{MASS:} masked
  sequence to sequence pre-training for language generation}.
\newblock In \emph{Proc. of ICML}, volume~97 of \emph{Proceedings of Machine
  Learning Research}, pages 5926--5936. {PMLR}.

\bibitem[{Srivastava et~al.(2022)Srivastava, Rastogi, Rao, Shoeb, Abid, Fisch,
  Brown, Santoro, Gupta, Garriga-Alonso et~al.}]{srivastava2022beyond}
Aarohi Srivastava, Abhinav Rastogi, Abhishek Rao, Abu Awal~Md Shoeb, Abubakar
  Abid, Adam Fisch, Adam~R Brown, Adam Santoro, Aditya Gupta, Adri{\`a}
  Garriga-Alonso, et~al. 2022.
\newblock \href {https://arxiv.org/abs/2206.04615} {Beyond the imitation game:
  Quantifying and extrapolating the capabilities of language models}.
\newblock \emph{ArXiv preprint}, abs/2206.04615.

\bibitem[{Studer et~al.(1998)Studer, Benjamins, and
  Fensel}]{studer1998knowledge}
Rudi Studer, V~Richard Benjamins, and Dieter Fensel. 1998.
\newblock Knowledge engineering: principles and methods.
\newblock \emph{Data \& knowledge engineering}, 25(1-2):161--197.

\bibitem[{Su et~al.(2022)Su, Kasai, Wu, Shi, Wang, Xin, Zhang, Ostendorf,
  Zettlemoyer, Smith et~al.}]{su2022selective}
Hongjin Su, Jungo Kasai, Chen~Henry Wu, Weijia Shi, Tianlu Wang, Jiayi Xin, Rui
  Zhang, Mari Ostendorf, Luke Zettlemoyer, Noah~A Smith, et~al. 2022.
\newblock \href {https://arxiv.org/abs/2209.01975} {Selective annotation makes
  language models better few-shot learners}.
\newblock \emph{ArXiv preprint}, abs/2209.01975.

\bibitem[{Sun et~al.(2022)Sun, Shao, Qian, Huang, and Qiu}]{sun2022black}
Tianxiang Sun, Yunfan Shao, Hong Qian, Xuanjing Huang, and Xipeng Qiu. 2022.
\newblock Black-box tuning for language-model-as-a-service.
\newblock In \emph{International Conference on Machine Learning}, pages
  20841--20855. PMLR.

\bibitem[{Sun et~al.(2019)Sun, Wang, Li, Feng, Chen, Zhang, Tian, Zhu, Tian,
  and Wu}]{sun2019ernie}
Yu~Sun, Shuohuan Wang, Yukun Li, Shikun Feng, Xuyi Chen, Han Zhang, Xin Tian,
  Danxiang Zhu, Hao Tian, and Hua Wu. 2019.
\newblock \href {https://arxiv.org/abs/1904.09223} {Ernie: Enhanced
  representation through knowledge integration}.
\newblock \emph{ArXiv preprint}, abs/1904.09223.

\bibitem[{Sung et~al.(2021)Sung, Lee, Yi, Jeon, Kim, and Kang}]{sung2021can}
Mujeen Sung, Jinhyuk Lee, Sean Yi, Minji Jeon, Sungdong Kim, and Jaewoo Kang.
  2021.
\newblock \href {https://doi.org/10.18653/v1/2021.emnlp-main.388} {Can language
  models be biomedical knowledge bases?}
\newblock In \emph{Proc. of EMNLP}, pages 4723--4734, Online and Punta Cana,
  Dominican Republic. Association for Computational Linguistics.

\bibitem[{Talmor et~al.(2020)Talmor, Elazar, Goldberg, and
  Berant}]{Talmor2020oLMpicsOnWL}
Alon Talmor, Yanai Elazar, Yoav Goldberg, and Jonathan Berant. 2020.
\newblock \href {https://doi.org/10.1162/tacl_a_00342} {o{LM}pics-on what
  language model pre-training captures}.
\newblock \emph{Transactions of the Association for Computational Linguistics},
  8:743--758.

\bibitem[{Tamborrino et~al.(2020)Tamborrino, Pellican{\`o}, Pannier, Voitot,
  and Naudin}]{tamborrinoPretrainingAlmostAll2020a}
Alexandre Tamborrino, Nicola Pellican{\`o}, Baptiste Pannier, Pascal Voitot,
  and Louise Naudin. 2020.
\newblock \href {https://doi.org/10.18653/v1/2020.acl-main.357} {Pre-training
  is (almost) all you need: An application to commonsense reasoning}.
\newblock In \emph{Proc. of ACL}, pages 3878--3887, Online. Association for
  Computational Linguistics.

\bibitem[{Tenney et~al.(2019)Tenney, Xia, Chen, Wang, Poliak, McCoy, Kim,
  Durme, Bowman, Das, and Pavlick}]{tenneyWhatYouLearn2019}
Ian Tenney, Patrick Xia, Berlin Chen, Alex Wang, Adam Poliak, R.~Thomas McCoy,
  Najoung Kim, Benjamin~Van Durme, Samuel~R. Bowman, Dipanjan Das, and Ellie
  Pavlick. 2019.
\newblock \href {https://openreview.net/forum?id=SJzSgnRcKX} {What do you learn
  from context? probing for sentence structure in contextualized word
  representations}.
\newblock In \emph{Proc. of ICLR}. OpenReview.net.

\bibitem[{Tian et~al.(2020)Tian, Gao, Xiao, Liu, He, Wu, Wang, and
  Wu}]{tian2020skep}
Hao Tian, Can Gao, Xinyan Xiao, Hao Liu, Bolei He, Hua Wu, Haifeng Wang, and
  Feng Wu. 2020.
\newblock \href {https://doi.org/10.18653/v1/2020.acl-main.374} {{SKEP}:
  Sentiment knowledge enhanced pre-training for sentiment analysis}.
\newblock In \emph{Proc. of ACL}, pages 4067--4076, Online. Association for
  Computational Linguistics.

\bibitem[{Vaswani et~al.(2017)Vaswani, Shazeer, Parmar, Uszkoreit, Jones,
  Gomez, Kaiser, and Polosukhin}]{vaswani2017attention}
Ashish Vaswani, Noam Shazeer, Niki Parmar, Jakob Uszkoreit, Llion Jones,
  Aidan~N. Gomez, Lukasz Kaiser, and Illia Polosukhin. 2017.
\newblock \href
  {https://proceedings.neurips.cc/paper/2017/hash/3f5ee243547dee91fbd053c1c4a845aa-Abstract.html}
  {Attention is all you need}.
\newblock In \emph{Advances in Neural Information Processing Systems 30: Annual
  Conference on Neural Information Processing Systems 2017, December 4-9, 2017,
  Long Beach, CA, {USA}}, pages 5998--6008.

\bibitem[{Vig et~al.(2020)Vig, Gehrmann, Belinkov, Qian, Nevo, Singer, and
  Shieber}]{Vig2020InvestigatingGB}
Jesse Vig, Sebastian Gehrmann, Yonatan Belinkov, Sharon Qian, Daniel Nevo,
  Yaron Singer, and Stuart~M. Shieber. 2020.
\newblock \href
  {https://proceedings.neurips.cc/paper/2020/hash/92650b2e92217715fe312e6fa7b90d82-Abstract.html}
  {Investigating gender bias in language models using causal mediation
  analysis}.
\newblock In \emph{Advances in Neural Information Processing Systems 33: Annual
  Conference on Neural Information Processing Systems 2020, NeurIPS 2020,
  December 6-12, 2020, virtual}.

\bibitem[{Voita and Titov(2020)}]{voita2020information}
Elena Voita and Ivan Titov. 2020.
\newblock \href {https://doi.org/10.18653/v1/2020.emnlp-main.14}
  {Information-theoretic probing with minimum description length}.
\newblock In \emph{Proc. of EMNLP}, pages 183--196, Online. Association for
  Computational Linguistics.

\bibitem[{von Oswald et~al.(2022)von Oswald, Niklasson, Randazzo, Sacramento,
  Mordvintsev, Zhmoginov, and Vladymyrov}]{von2022transformers}
Johannes von Oswald, Eyvind Niklasson, Ettore Randazzo, Jo{\~a}o Sacramento,
  Alexander Mordvintsev, Andrey Zhmoginov, and Max Vladymyrov. 2022.
\newblock \href {https://arxiv.org/abs/2212.07677} {Transformers learn
  in-context by gradient descent}.
\newblock \emph{ArXiv preprint}, abs/2212.07677.

\bibitem[{Vrande\v{c}i\'{c} and Kr\"{o}tzsch(2014)}]{10.1145/2629489}
Denny Vrande\v{c}i\'{c} and Markus Kr\"{o}tzsch. 2014.
\newblock \href {https://doi.org/10.1145/2629489} {Wikidata: A free
  collaborative knowledgebase}.
\newblock \emph{Commun. ACM}, 57(10):78–85.

\bibitem[{Wallace et~al.(2019)Wallace, Wang, Li, Singh, and
  Gardner}]{wallaceNLPModelsKnow2019}
Eric Wallace, Yizhong Wang, Sujian Li, Sameer Singh, and Matt Gardner. 2019.
\newblock \href {https://doi.org/10.18653/v1/D19-1534} {Do {NLP} models know
  numbers? probing numeracy in embeddings}.
\newblock In \emph{Proc. of EMNLP}, pages 5307--5315, Hong Kong, China.
  Association for Computational Linguistics.

\bibitem[{Wallat et~al.(2020)Wallat, Singh, and Anand}]{wallat2021bertnesia}
Jonas Wallat, Jaspreet Singh, and Avishek Anand. 2020.
\newblock \href {https://doi.org/10.18653/v1/2020.blackboxnlp-1.17}
  {{BERT}nesia: Investigating the capture and forgetting of knowledge in
  {BERT}}.
\newblock In \emph{Proceedings of the Third BlackboxNLP Workshop on Analyzing
  and Interpreting Neural Networks for NLP}, pages 174--183, Online.
  Association for Computational Linguistics.

\bibitem[{Wang et~al.(2020)Wang, Liu, and Song}]{wang2020language}
Chenguang Wang, Xiao Liu, and Dawn Song. 2020.
\newblock \href {https://arxiv.org/abs/2010.11967} {Language models are open
  knowledge graphs}.
\newblock \emph{ArXiv preprint}, abs/2010.11967.

\bibitem[{Wang et~al.(2021{\natexlab{a}})Wang, Tang, Duan, Wei, Huang, Ji, Cao,
  Jiang, and Zhou}]{wang2020k}
Ruize Wang, Duyu Tang, Nan Duan, Zhongyu Wei, Xuanjing Huang, Jianshu Ji,
  Guihong Cao, Daxin Jiang, and Ming Zhou. 2021{\natexlab{a}}.
\newblock \href {https://doi.org/10.18653/v1/2021.findings-acl.121}
  {{K-Adapter}: {I}nfusing {K}nowledge into {P}re-{T}rained {M}odels with
  {A}dapters}.
\newblock In \emph{Findings of the Association for Computational Linguistics:
  ACL-IJCNLP 2021}, pages 1405--1418, Online. Association for Computational
  Linguistics.

\bibitem[{Wang et~al.(2021{\natexlab{b}})Wang, Gao, Zhu, Zhang, Liu, Li, and
  Tang}]{wang2021kepler}
Xiaozhi Wang, Tianyu Gao, Zhaocheng Zhu, Zhengyan Zhang, Zhiyuan Liu, Juanzi
  Li, and Jian Tang. 2021{\natexlab{b}}.
\newblock \href {https://doi.org/10.1162/tacl_a_00360} {{KEPLER}: A unified
  model for knowledge embedding and pre-trained language representation}.
\newblock \emph{Transactions of the Association for Computational Linguistics},
  9:176--194.

\bibitem[{Warstadt et~al.(2019)Warstadt, Cao, Grosu, Peng, Blix, Nie, Alsop,
  Bordia, Liu, Parrish, Wang, Phang, Mohananey, Htut, Jeretic, and
  Bowman}]{warstadtInvestigatingBERTKnowledge2019}
Alex Warstadt, Yu~Cao, Ioana Grosu, Wei Peng, Hagen Blix, Yining Nie, Anna
  Alsop, Shikha Bordia, Haokun Liu, Alicia Parrish, Sheng-Fu Wang, Jason Phang,
  Anhad Mohananey, Phu~Mon Htut, Paloma Jeretic, and Samuel~R. Bowman. 2019.
\newblock \href {https://doi.org/10.18653/v1/D19-1286} {Investigating
  {BERT}{'}s knowledge of language: Five analysis methods with {NPI}s}.
\newblock In \emph{Proc. of EMNLP}, pages 2877--2887, Hong Kong, China.
  Association for Computational Linguistics.

\bibitem[{Warstadt et~al.(2020)Warstadt, Parrish, Liu, Mohananey, Peng, Wang,
  and Bowman}]{warstadt-etal-2020-blimp-benchmark}
Alex Warstadt, Alicia Parrish, Haokun Liu, Anhad Mohananey, Wei Peng, Sheng-Fu
  Wang, and Samuel~R. Bowman. 2020.
\newblock \href {https://doi.org/10.1162/tacl_a_00321} {{BL}i{MP}: The
  benchmark of linguistic minimal pairs for {E}nglish}.
\newblock \emph{Transactions of the Association for Computational Linguistics},
  8:377--392.

\bibitem[{Wei et~al.(2021{\natexlab{a}})Wei, Bosma, Zhao, Guu, Yu, Lester, Du,
  Dai, and Le}]{wei2021finetuned}
Jason Wei, Maarten Bosma, Vincent~Y Zhao, Kelvin Guu, Adams~Wei Yu, Brian
  Lester, Nan Du, Andrew~M Dai, and Quoc~V Le. 2021{\natexlab{a}}.
\newblock \href {https://arxiv.org/abs/2109.01652} {Finetuned language models
  are zero-shot learners}.
\newblock \emph{ArXiv preprint}, abs/2109.01652.

\bibitem[{Wei et~al.(2022)Wei, Wang, Schuurmans, Bosma, Chi, Le, and
  Zhou}]{wei2022chain}
Jason Wei, Xuezhi Wang, Dale Schuurmans, Maarten Bosma, Ed~Chi, Quoc Le, and
  Denny Zhou. 2022.
\newblock \href {https://arxiv.org/abs/2201.11903} {Chain of thought prompting
  elicits reasoning in large language models}.
\newblock \emph{ArXiv preprint}, abs/2201.11903.

\bibitem[{Wei et~al.(2021{\natexlab{b}})Wei, Wang, Zhang, Bhatia, and
  Arnold}]{wei2021knowledge}
Xiaokai Wei, Shen Wang, Dejiao Zhang, Parminder Bhatia, and Andrew Arnold.
  2021{\natexlab{b}}.
\newblock \href {https://arxiv.org/abs/2110.08455} {Knowledge enhanced
  pretrained language models: A compreshensive survey}.
\newblock \emph{ArXiv preprint}, abs/2110.08455.

\bibitem[{Wu et~al.(2020)Wu, Chen, Kao, and Liu}]{wu2020perturbed}
Zhiyong Wu, Yun Chen, Ben Kao, and Qun Liu. 2020.
\newblock \href {https://doi.org/10.18653/v1/2020.acl-main.383} {Perturbed
  masking: Parameter-free probing for analyzing and interpreting {BERT}}.
\newblock In \emph{Proc. of ACL}, pages 4166--4176, Online. Association for
  Computational Linguistics.

\bibitem[{Xiong et~al.(2020)Xiong, Du, Wang, and
  Stoyanov}]{xiong2019pretrained}
Wenhan Xiong, Jingfei Du, William~Yang Wang, and Veselin Stoyanov. 2020.
\newblock \href {https://openreview.net/forum?id=BJlzm64tDH} {Pretrained
  encyclopedia: Weakly supervised knowledge-pretrained language model}.
\newblock In \emph{Proc. of ICLR}. OpenReview.net.

\bibitem[{Yaghoobzadeh et~al.(2019)Yaghoobzadeh, Kann, Hazen, Agirre, and
  Sch{\"u}tze}]{yaghoobzadeh2019probing}
Yadollah Yaghoobzadeh, Katharina Kann, T.~J. Hazen, Eneko Agirre, and Hinrich
  Sch{\"u}tze. 2019.
\newblock \href {https://doi.org/10.18653/v1/P19-1574} {Probing for semantic
  classes: Diagnosing the meaning content of word embeddings}.
\newblock In \emph{Proc. of ACL}, pages 5740--5753, Florence, Italy.
  Association for Computational Linguistics.

\bibitem[{Yamada et~al.(2020)Yamada, Asai, Shindo, Takeda, and
  Matsumoto}]{yamada2020luke}
Ikuya Yamada, Akari Asai, Hiroyuki Shindo, Hideaki Takeda, and Yuji Matsumoto.
  2020.
\newblock \href {https://doi.org/10.18653/v1/2020.emnlp-main.523} {{LUKE}: Deep
  contextualized entity representations with entity-aware self-attention}.
\newblock In \emph{Proc. of EMNLP}, pages 6442--6454, Online. Association for
  Computational Linguistics.

\bibitem[{Yang et~al.(2021)Yang, Xiao, Shen, Jiang, Hu, Zhang, and
  Peng}]{yang2021survey}
Jian Yang, Gang Xiao, Yulong Shen, Wei Jiang, Xinyu Hu, Ying Zhang, and Jinghui
  Peng. 2021.
\newblock \href {https://arxiv.org/abs/2110.00269} {A survey of knowledge
  enhanced pre-trained models}.
\newblock \emph{ArXiv preprint}, abs/2110.00269.

\bibitem[{Yasunaga et~al.(2022)Yasunaga, Aghajanyan, Shi, James, Leskovec,
  Liang, Lewis, Zettlemoyer, and Yih}]{DBLP:journals/corr/abs-2211-12561}
Michihiro Yasunaga, Armen Aghajanyan, Weijia Shi, Rich James, Jure Leskovec,
  Percy Liang, Mike Lewis, Luke Zettlemoyer, and Wen{-}tau Yih. 2022.
\newblock \href {https://doi.org/10.48550/arXiv.2211.12561}
  {Retrieval-augmented multimodal language modeling}.
\newblock \emph{CoRR}, abs/2211.12561.

\bibitem[{Ye et~al.(2019)Ye, Chen, Wang, and Ling}]{ye2019align}
Zhi-Xiu Ye, Qian Chen, Wen Wang, and Zhen-Hua Ling. 2019.
\newblock \href {https://arxiv.org/abs/1908.06725} {Align, mask and select: A
  simple method for incorporating commonsense knowledge into language
  representation models}.
\newblock \emph{ArXiv preprint}, abs/1908.06725.

\bibitem[{Yin et~al.(2022)Yin, Dong, Cheng, Liu, Chang, Wei, and
  Gao}]{yin2022survey}
Da~Yin, Li~Dong, Hao Cheng, Xiaodong Liu, Kai-Wei Chang, Furu Wei, and Jianfeng
  Gao. 2022.
\newblock \href {https://arxiv.org/abs/2202.08772} {A survey of
  knowledge-intensive nlp with pre-trained language models}.
\newblock \emph{ArXiv preprint}, abs/2202.08772.

\bibitem[{Yu et~al.(2022)Yu, Iter, Wang, Xu, Ju, Sanyal, Zhu, Zeng, and
  Jiang}]{yu2022generate}
Wenhao Yu, Dan Iter, Shuohang Wang, Yichong Xu, Mingxuan Ju, Soumya Sanyal,
  Chenguang Zhu, Michael Zeng, and Meng Jiang. 2022.
\newblock \href {https://arxiv.org/abs/2209.10063} {Generate rather than
  retrieve: Large language models are strong context generators}.
\newblock \emph{ArXiv preprint}, abs/2209.10063.

\bibitem[{Zhang et~al.(2022{\natexlab{a}})Zhang, Zhang, Zhang, and
  Yang}]{zhang2022robustness}
Hongxin Zhang, Yanzhe Zhang, Ruiyi Zhang, and Diyi Yang. 2022{\natexlab{a}}.
\newblock \href {https://arxiv.org/abs/2210.10693} {Robustness of
  demonstration-based learning under limited data scenario}.
\newblock \emph{ArXiv preprint}, abs/2210.10693.

\bibitem[{Zhang et~al.(2022{\natexlab{b}})Zhang, Roller, Goyal, Artetxe, Chen,
  Chen, Dewan, Diab, Li, Lin, Mihaylov, Ott, Shleifer, Shuster, Simig, Koura,
  Sridhar, Wang, and Zettlemoyer}]{zhang2022opt}
Susan Zhang, Stephen Roller, Naman Goyal, Mikel Artetxe, Moya Chen, Shuohui
  Chen, Christopher Dewan, Mona Diab, Xian Li, Xi~Victoria Lin, Todor Mihaylov,
  Myle Ott, Sam Shleifer, Kurt Shuster, Daniel Simig, Punit~Singh Koura, Anjali
  Sridhar, Tianlu Wang, and Luke Zettlemoyer. 2022{\natexlab{b}}.
\newblock \href {http://arxiv.org/abs/2205.01068} {Opt: Open pre-trained
  transformer language models}.

\bibitem[{Zhang et~al.(2019)Zhang, Han, Liu, Jiang, Sun, and
  Liu}]{zhang2019ernie}
Zhengyan Zhang, Xu~Han, Zhiyuan Liu, Xin Jiang, Maosong Sun, and Qun Liu. 2019.
\newblock \href {https://doi.org/10.18653/v1/P19-1139} {{ERNIE}: Enhanced
  language representation with informative entities}.
\newblock In \emph{Proc. of ACL}, pages 1441--1451, Florence, Italy.
  Association for Computational Linguistics.

\bibitem[{Zhao et~al.(2021)Zhao, Wallace, Feng, Klein, and
  Singh}]{zhao2021calibrate}
Zihao Zhao, Eric Wallace, Shi Feng, Dan Klein, and Sameer Singh. 2021.
\newblock \href {http://proceedings.mlr.press/v139/zhao21c.html} {Calibrate
  before use: Improving few-shot performance of language models}.
\newblock In \emph{Proc. of ICML}, volume 139 of \emph{Proceedings of Machine
  Learning Research}, pages 12697--12706. {PMLR}.

\bibitem[{Zhong et~al.(2021)Zhong, Friedman, and Chen}]{zhong2021factual}
Zexuan Zhong, Dan Friedman, and Danqi Chen. 2021.
\newblock \href {https://doi.org/10.18653/v1/2021.naacl-main.398} {Factual
  probing is [{MASK}]: Learning vs. learning to recall}.
\newblock In \emph{Proceedings of the 2021 Conference of the North American
  Chapter of the Association for Computational Linguistics: Human Language
  Technologies}, pages 5017--5033, Online. Association for Computational
  Linguistics.

\bibitem[{Zhou et~al.(2022)Zhou, Sch{\"a}rli, Hou, Wei, Scales, Wang,
  Schuurmans, Bousquet, Le, and Chi}]{zhou2022least}
Denny Zhou, Nathanael Sch{\"a}rli, Le~Hou, Jason Wei, Nathan Scales, Xuezhi
  Wang, Dale Schuurmans, Olivier Bousquet, Quoc Le, and Ed~Chi. 2022.
\newblock \href {https://arxiv.org/abs/2205.10625} {Least-to-most prompting
  enables complex reasoning in large language models}.
\newblock \emph{ArXiv preprint}, abs/2205.10625.

\bibitem[{Zhou et~al.(2019)Zhou, Zhang, Zhao, and Zhang}]{zhou2019limit}
Junru Zhou, Zhuosheng Zhang, Hai Zhao, and Shuailiang Zhang. 2019.
\newblock \href {https://arxiv.org/abs/1910.14296} {Limit-bert: Linguistic
  informed multi-task bert}.
\newblock \emph{ArXiv preprint}, abs/1910.14296.

\bibitem[{Zhou et~al.(2020{\natexlab{a}})Zhou, Zhang, Cui, and
  Huang}]{Zhou2020EvaluatingCI}
Xuhui Zhou, Yue Zhang, Leyang Cui, and Dandan Huang. 2020{\natexlab{a}}.
\newblock \href {https://aaai.org/ojs/index.php/AAAI/article/view/6523}
  {Evaluating commonsense in pre-trained language models}.
\newblock In \emph{The Thirty-Fourth {AAAI} Conference on Artificial
  Intelligence, {AAAI} 2020, The Thirty-Second Innovative Applications of
  Artificial Intelligence Conference, {IAAI} 2020, The Tenth {AAAI} Symposium
  on Educational Advances in Artificial Intelligence, {EAAI} 2020, New York,
  NY, USA, February 7-12, 2020}, pages 9733--9740. {AAAI} Press.

\bibitem[{Zhou et~al.(2020{\natexlab{b}})Zhou, Zhang, Cui, and
  Huang}]{zhouEvaluatingCommonsensePretrained2019}
Xuhui Zhou, Yue Zhang, Leyang Cui, and Dandan Huang. 2020{\natexlab{b}}.
\newblock \href {https://aaai.org/ojs/index.php/AAAI/article/view/6523}
  {Evaluating commonsense in pre-trained language models}.
\newblock In \emph{The Thirty-Fourth {AAAI} Conference on Artificial
  Intelligence, {AAAI} 2020, The Thirty-Second Innovative Applications of
  Artificial Intelligence Conference, {IAAI} 2020, The Tenth {AAAI} Symposium
  on Educational Advances in Artificial Intelligence, {EAAI} 2020, New York,
  NY, USA, February 7-12, 2020}, pages 9733--9740. {AAAI} Press.

\bibitem[{Zhou and Srikumar(2021{\natexlab{a}})}]{zhou2021directprobe}
Yichu Zhou and Vivek Srikumar. 2021{\natexlab{a}}.
\newblock \href {https://doi.org/10.18653/v1/2021.naacl-main.401}
  {{D}irect{P}robe: Studying representations without classifiers}.
\newblock In \emph{Proceedings of the 2021 Conference of the North American
  Chapter of the Association for Computational Linguistics: Human Language
  Technologies}, pages 5070--5083, Online. Association for Computational
  Linguistics.

\bibitem[{Zhou and
  Srikumar(2021{\natexlab{b}})}]{zhou-srikumar-2021-directprobe}
Yichu Zhou and Vivek Srikumar. 2021{\natexlab{b}}.
\newblock \href {https://doi.org/10.18653/v1/2021.naacl-main.401}
  {{D}irect{P}robe: Studying representations without classifiers}.
\newblock In \emph{Proceedings of the 2021 Conference of the North American
  Chapter of the Association for Computational Linguistics: Human Language
  Technologies}, pages 5070--5083, Online. Association for Computational
  Linguistics.

\bibitem[{Zhu et~al.(2020)Zhu, Rawat, Zaheer, Bhojanapalli, Li, Yu, and
  Kumar}]{zhu2020modifying}
Chen Zhu, Ankit~Singh Rawat, Manzil Zaheer, Srinadh Bhojanapalli, Daliang Li,
  Felix Yu, and Sanjiv Kumar. 2020.
\newblock \href {https://arxiv.org/abs/2012.00363} {Modifying memories in
  transformer models}.
\newblock \emph{ArXiv preprint}, abs/2012.00363.

\bibitem[{Zimbardo and Ruch(1975)}]{zimbardo1975psychology}
Philip~G Zimbardo and Floyd~L Ruch. 1975.
\newblock Psychology and life.

\end{thebibliography}
